\documentclass[lettersize,journal]{IEEEtran}
\usepackage{amsmath,amsfonts}
\usepackage{algorithmic}
\usepackage{algorithm}
\usepackage{array}
\usepackage[caption=false,font=normalsize,labelfont=sf,textfont=sf]{subfig}
\usepackage{textcomp}
\usepackage{stfloats}
\usepackage{url}
\usepackage{verbatim}
\usepackage{graphicx}
\usepackage{cite}
\usepackage{graphics} 
\usepackage{epsfig} 
\usepackage{mathptmx} 
\usepackage{times} 
\usepackage{amssymb}  
\usepackage[T1]{fontenc}
\usepackage{xcolor}
\usepackage[all]{nowidow}
\usepackage{tikz,hyperref}
\usepackage{microtype}
\usepackage{makecell}
\usepackage{comment}
\usepackage{multirow}
\usepackage{arydshln}

\begin{document}

\title{Predicting Pedestrian Crossing Behavior in Germany and Japan: Insights into Model Transferability}

\author{Chi Zhang,
Janis Sprenger, 
Zhongjun Ni, 
and Christian Berger
\thanks{This research is partially supported by the research project ``SHAPE-IT – Supporting the Interaction of Humans and Automated Vehicles: Preparing for the Environment of Tomorrow’’ which has received funding from the European Union’s Horizon 2020 research and innovation programme under the Marie Skłodowska-Curie grant agreement 860410, partially supported by the Swedish Foundation for Strategic Research (SSF), Grant Number FUS21-0004 (SAICOM), and partially supported by the German Ministry for Research and Education (BMBF) in the project MOMENTUM (01IW22001).}

\thanks{Chi Zhang and Christian Berger are with the Department of Computer Science and Engineering, University of Gothenburg, Gothenburg, Sweden (email: chi.zhang@gu.se, christian.berger@gu.se).}%
\thanks{Janis Sprenger is with the German Research Center for Artificial Intelligence (DFKI), Saarland Informatics Campus, Germany (email: janis.sprenger@dfki.de).}
\thanks{Zhongjun Ni is with the Department of Science and Technology, Linköping University, Campus Norrk{\"o}ping, Norrk{\"o}ping, Sweden (email: zhongjun.ni@liu.se).}
\thanks{\textit{Corresponding author: Christian Berger.}}
}

\maketitle

\begin{abstract}
Predicting pedestrian crossing behavior is important for intelligent traffic systems to avoid pedestrian-vehicle collisions. Most existing pedestrian crossing behavior models are trained and evaluated on datasets collected from a single country, overlooking differences between countries. To address this gap, we compared pedestrian road-crossing behavior at unsignalized crossings in Germany and Japan. We presented four types of machine learning models to predict gap selection behavior, zebra crossing usage, and their trajectories using simulator data collected from both countries. When comparing the differences between countries, pedestrians from the study conducted in Japan are more cautious, selecting larger gaps compared to those in Germany. We evaluate and analyze model transferability. Our results show that neural networks outperform other machine learning models in predicting gap selection and zebra crossing usage, while random forest models perform best on trajectory prediction tasks, demonstrating strong performance and transferability. We develop a transferable model using an unsupervised clustering method, which improves prediction accuracy for gap selection and trajectory prediction. These findings provide a deeper understanding of pedestrian crossing behaviors in different countries and offer valuable insights into model transferability.
\end{abstract}


\section{Introduction}
Intelligent driving systems and smart traffic infrastructures are developed to enable safer and smarter travel. As pedestrians are vulnerable, understanding and predicting their behavior is crucial, especially at unsignalized crossings without traffic lights, where pedestrians constantly interact with traffic due to the ambiguity of the right-of-way. Predicting pedestrian behavior is challenging, as many factors influence pedestrian behavior ~\cite{zhang2023pedestrian,Rasouli2019Autonomous}. Previous research studies focused on tasks including crossing intention prediction~\cite{Rasouli2017they, Fang2018, Rasouli2019PIE, Chaabane2020, yang2021crossing, zhang2021pedestrian} and trajectory prediction~\cite{alahi2016social,gupta2018social,mohamed2020social,zhang2021social,zhang2022learning,zhang2023spatial}, but many of them did not explicitly consider pedestrian-vehicle interactions.

When pedestrians interact with vehicles at unsignalized crossings, intelligent traffic systems should be able to predict their behavior, such as the time gap they will choose to cross, whether they will use zebra crossings, and their crossing trajectories. Zhang et al.~\cite{zhang2024predicting} has predicted and analyzed pedestrian gap selection and zebra crossing usage using data collected by a simulator study conducted in Germany. Their study provides insights into the key factors that influence pedestrian behavior when interacting with multiple vehicles. 

However, Zhang et al.'s study~\cite{zhang2024predicting} and most other existing models only focused on single datasets, neglecting the differences between countries. An intelligent traffic system should be functional and safe in different countries. In other words, the model trained in one country should be transferable to other countries. Therefore, in this study, we investigate and compare the differences between Germany and Japan, providing insights for bridging differences between countries and improving model transferability.

Simulator studies have been widely used by researchers (e.g., \cite{zhang2023cross, sprenger2023cross, zhang2024predicting}) to investigate pedestrian-vehicle interactions, as it is safe and avoids real pedestrian-vehicle conflicts in near-crash events. In this context, the term ``pedestrian-vehicle interactions'' refers to the exchange of actions and reactions between pedestrians and vehicles that influence each other’s behavior during encounters. These interactions can be \textit{explicitly} represented and modeled by measurable variables such as time to arrival (TTA) or time gap, pedestrian waiting time, pedestrian walking speed, and vehicle speed. In this study, we use simulator data collected in a virtual reality (VR) simulator by Sprenger et al.~\cite{sprenger2023cross} from both Germany and Japan. The head-mounted VR systems ensure a consistent and controlled environment across different countries, reducing the impact of irrelevant factors and allowing us to focus on behavioral differences between the two countries.

This study extends Zhang et al.'s research~\cite{zhang2024predicting}, which explored scenarios where pedestrians interact with multiple vehicles in traffic flow using machine learning models. 
Compared to their research that only focused on data collected from Germany, we investigate the differences between data collected from Germany and Japan, and explore the transferability of these models. Furthermore, we expand the scope to include pedestrian crossing trajectories in addition to predicting gap selection behavior and zebra crossing usage.
Particularly, for each prediction task, our research questions are:

\begin{enumerate}
    \item [RQ1] What are the similarities and differences in pedestrian crossing behavior observed between studies conducted in Germany and Japan?
    \item [RQ2] How transferable are models trained on data collected from Germany and Japan when tested on the other dataset?
    \item [RQ3] How do unsupervised learning clustering methods enhance performance in building transferable models?
\end{enumerate}

The main contributions of this study include:
\begin{itemize}
    \item Proposing and evaluating machine learning models to predict pedestrian gap selection behavior, zebra crossing usage, and crossing trajectories in Germany and Japan, comparing behavioral patterns between the data collected from both countries.
    
    \item Evaluating the transferability of models trained on the data collected from Germany and Japan to each other.

    \item Proposing and comparing methods based on unsupervised learning-clustering to enhance model transferability and performance on datasets collected from Germany and Japan.
\end{itemize}

\section{Related Work}

In this section, we review related studies, focusing on three key areas: pedestrian crossing behavior prediction, analysis of pedestrian crossing behavior, and differences in pedestrian behavior between countries.

\subsection{Pedestrian Crossing Behavior Prediction}
Studies such as~\cite{alahi2016social,gupta2018social,mohamed2020social,zhang2021social,zhang2022learning,zhang2023spatial,Rasouli2019PIE} focused on pedestrian trajectory prediction, using past spatial and temporal information to predict future movements. These methods typically use a 3.2 s time window for observation and output a 4.8 s prediction.
However, they only rely on pedestrians' past trajectories and do not explicitly consider pedestrian-vehicle interaction factors, failing to capture the complexities of crossing decisions that can extend beyond 8 s.
To bridge this gap, this study considers interaction factors to predict pedestrians' crossing paths without relying on their past trajectories.

Some studies focused on pedestrian crossing intentions and applied deep learning on naturalistic data to determine whether a pedestrian intends to cross. Studies such as~\cite{Rasouli2017they,Fang2018,Chaabane2020,zhang2021pedestrian} mainly considered only pedestrian-related features, including pedestrian appearance, their surroundings, and pedestrian skeleton information. 
Other studies implicitly considered pedestrian-vehicle interactions by using relative pedestrian-vehicle distances~\cite{yang2021crossing} or their trajectories~\cite{Rasouli2019PIE} as inputs to deep learning networks, or by modeling interactions through graph convolutional networks~\cite{zhou2023pedestrian,ling2023stma}.
However, these approaches rely on implicit interaction models, which lack explainability. An explicitly defined, interpretable model of pedestrian-vehicle interactions is often missing due to the complexity of controlling the environment in naturalistic data and the difficulty in collecting near-crash scenarios for safety reasons.

Studies~\cite{Volz2015, Volz2016, Zhang2020Research, zhang2023cross} investigated pedestrian-vehicle interaction, focusing on scenarios where pedestrians interact with a single vehicle, considering factors such as time to arrival, position, velocity, pedestrian waiting time, and personal traits. These studies neglected the complex situations where pedestrians interact with multiple vehicles, which is common in real-world environments. Simulator studies provide a controlled environment to ensure pedestrian safety and collect detailed interaction information, avoiding real-world near-crash scenarios and are utilized by many researchers. Jayaraman et al.~\cite{jayaraman2020analysis} analyzed and predicted gap acceptance behavior at zebra crossings, considering variables like waiting time and vehicle speed. Zhang et al.~\cite{zhang2024predicting} used gap-related information to predict which gaps pedestrians would select for crossing and whether they would use zebra crossings.

However, existing models are typically trained and tested on single datasets, disregarding differences between countries, and lacking consideration for their transferability to new scenarios and countries. Our study aims to fill these gaps by evaluating model transferability that considers differences between countries and using cluster methods to build transferable models.

\subsection{Analysis of Pedestrian Crossing Behavior}

Pedestrian crossing behavior is influenced by various factors~\cite{zhang2023pedestrian, Rasouli2019Autonomous}. Pedestrian-related factors such as age, gender, and personality traits are explored in studies~\cite{cloutier2017outta,wang2022effect,rosenbloom2006sensation, gorrini2018observation, Rasouli2019Autonomous,velasco2021will, Kalantari2022Who}. Vehicle-related factors, such as vehicle speed, have also been considered in studies~\cite{Volz2016, theofilatos2021cross, yannis2013pedestrian}. Additionally, pedestrian-vehicle interaction plays a crucial role, as discussed in studies~\cite{gorrini2018observation, theofilatos2021cross, velasco2021will, Kalantari2022Who, zhang2023cross}. The time gap, an important factor for crossing decisions, has been explored in several studies~\cite{gorrini2018observation, theofilatos2021cross, velasco2021will, Kalantari2022Who, zhang2023cross}. Pedestrian waiting time has also been investigated~\cite{Kalantari2022Who, theofilatos2021cross, zhang2023cross, yannis2013pedestrian}.

However, these studies often modeled the relationship between these factors and crossing decisions without predicting specific outcomes. For example, Velasco et al.~\cite{velasco2021will} found that pedestrians are more likely to cross at larger gaps, while Gorrini et al.~\cite{gorrini2018observation} calculated accepted gaps from a statistical perspective. Yannis et al.~\cite{yannis2013pedestrian} modeled traffic gaps and crossing decisions at mid-block crossings, considering factors such as waiting time, vehicle speed, and group behavior, but their approach was limited to linear relationships and did not propose predictive models. Few studies focus on predicting the specific time gap a pedestrian would select and accept for crossing or whether a pedestrian would use the zebra crossings.

To address these research gaps, we consider pedestrians' interactions with multiple vehicles and predict interaction outcomes. For non-zebra crossing scenarios, we predict the time gap pedestrians select and accept when interacting with multiple vehicles, analyzing factors such as waiting time, pedestrian walking speed, and missed car gaps. For zebra crossings, we predict if pedestrians use the zebra crossing and analyze the factors influencing their choices.

\subsection{Pedestrian Behavior Differences between Countries}

Existing cross-country studies have explored behavioral differences. Hell et al.~\cite{hell2021pedestrian} provided a literature review on behavioral differences in risk avoidance, compliance, gap acceptance, and walking velocity between Germany and Japan. Following this, Sprenger et al.~\cite{sprenger2023cross} statistically compared pedestrian crossing velocity, gap sizes, following behavior, and zebra crossing usage in both countries based on a virtual reality simulator study conducted in both countries.
Solmazer et al.~\cite{solmazer2020cross} compared behaviors across Estonia, Greece, Kosovo, Russia, and Turkey, revealing that context or country-specific factors influence pedestrian behavior. Pele et al.~\cite{pele2017cultural, pele2019decision} analyzed pedestrian behavior differences between France and Japan, using agent-based modeling to uncover cognitive mechanisms and behavioral differences. Kaminka et al.~\cite{kaminka2018simulating} explicitly modeled cultural information in their simulation of pedestrian behaviors in Iraq, Israel, England, Canada, and France, highlighting movement differences and proposing mixed-country scenarios.

However, previous studies mainly compared behaviors across countries without building predictive models~\cite{hell2021pedestrian, sprenger2023cross, pele2017cultural}. Those that built models focused on specific factors~\cite{solmazer2020cross, pele2019decision,kaminka2018simulating} without developing transferable models, and few utilized machine learning methods.

To bridge this gap, our research compares differences between countries in predicting pedestrian crossing behavior and investigates the transferability of models trained on data from different countries. After comparing the differences between countries, we also used cluster methods to improve the performance and transferability.

We summarize research directions, main contributions, and research gaps of related studies in predicting and analyzing pedestrian crossing behavior in Table~\ref{tab:summary_literature}.

\begin{table}[tbh]
\centering
\caption{Summary of related work}
\label{tab:summary_literature}
\begin{tabular}{m{1.1cm}|m{2.8cm}|m{3.7cm}}
\hline
Research Direction & Main contributions & Research gaps \\
\hline
Behavior Prediction & Trajectory prediction~\cite{alahi2016social,gupta2018social,mohamed2020social,zhang2021social,zhang2022learning,zhang2023spatial,Rasouli2019PIE}; Crossing intention prediction~\cite{Rasouli2017they,Fang2018,Chaabane2020,yang2021crossing,zhang2021pedestrian, Rasouli2019PIE}; Pedestrian-vehicle interaction~\cite{Volz2015, Volz2016, Zhang2020Research, zhang2023cross, jayaraman2020analysis,zhang2024predicting}
 & Relied on past trajectories and did not explicitly consider interactions with vehicles; Trained and tested on single datasets without considering the difference between countries. \\ \hdashline

Behavior Analysis & 
Explored various factors influencing behavior using statistical analysis~\cite{cloutier2017outta, gorrini2018observation, Rasouli2019Autonomous, velasco2021will, Kalantari2022Who,rosenbloom2006sensation, wang2022effect, Volz2016, theofilatos2021cross, yannis2013pedestrian, zhang2023cross}  & Did not predict specific outcomes such as time gap selection, mainly used linear modeling approaches that limit predictive capabilities, and focused on single datasets without considering the difference between countries. \\ \hdashline

Cross-country Study & 
Explored cross-country pedestrian behavioral differences~\cite{hell2021pedestrian, sprenger2023cross, pele2017cultural,solmazer2020cross, pele2019decision,kaminka2018simulating} & 
Lacked development of transferable predictive models, and mainly focused on statistical comparison without machine learning \\

\hline
\end{tabular}
\end{table}

\section{Methodology}
\subsection{Data Collection}
\label{sec:data_collection}

\begin{figure*}[htb]
    \centering
    \subfloat[Baseline environment and virtual pedestrians]{\includegraphics[page=6,width=0.47\textwidth,trim=20 0 20 0,clip]{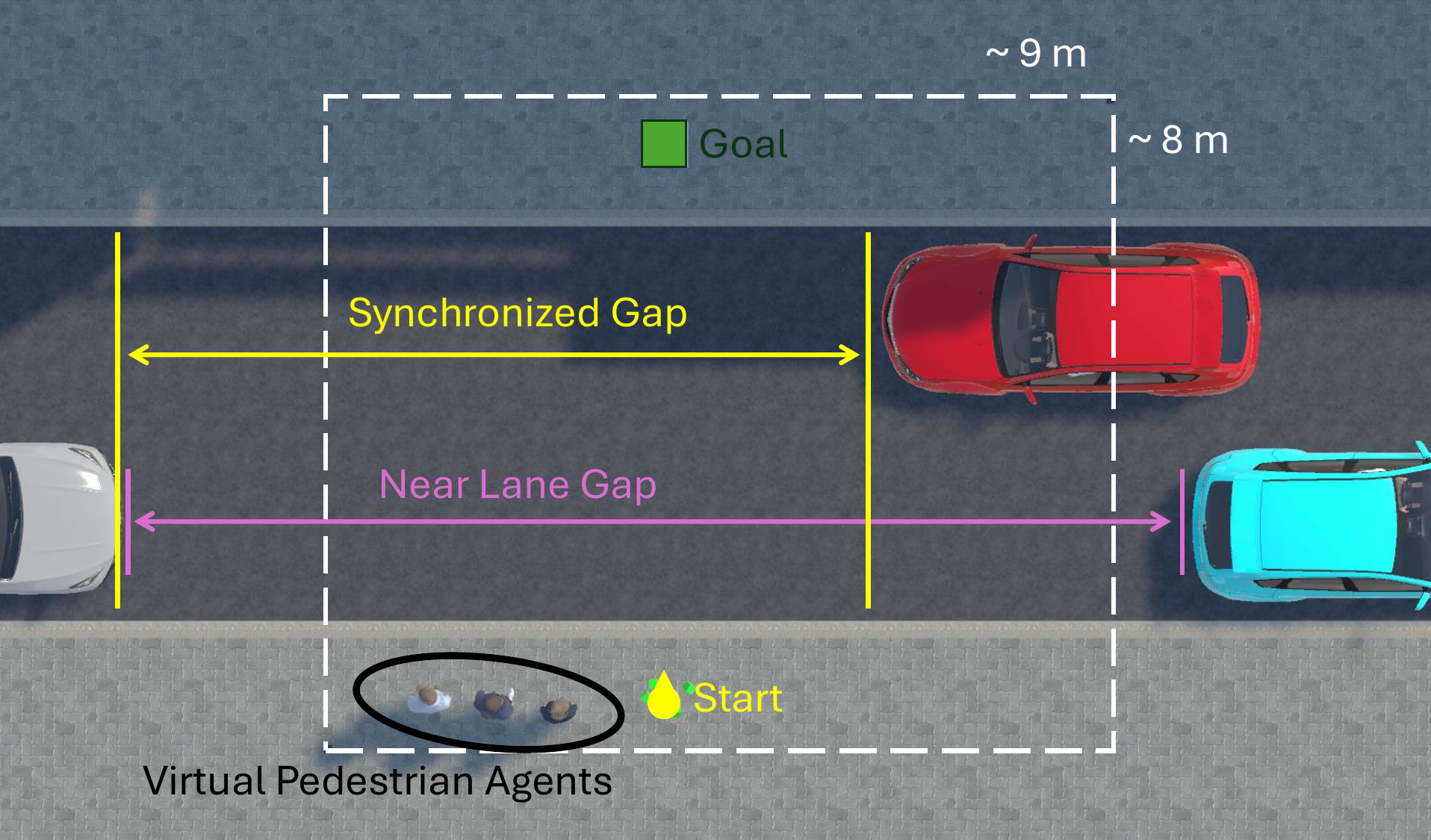}
        \label{fig:schematicPedestrians}}
    \hfill
    \subfloat[Zebra environment]{\includegraphics[page=3,width=0.47\textwidth,trim=20 0 20 0,clip]{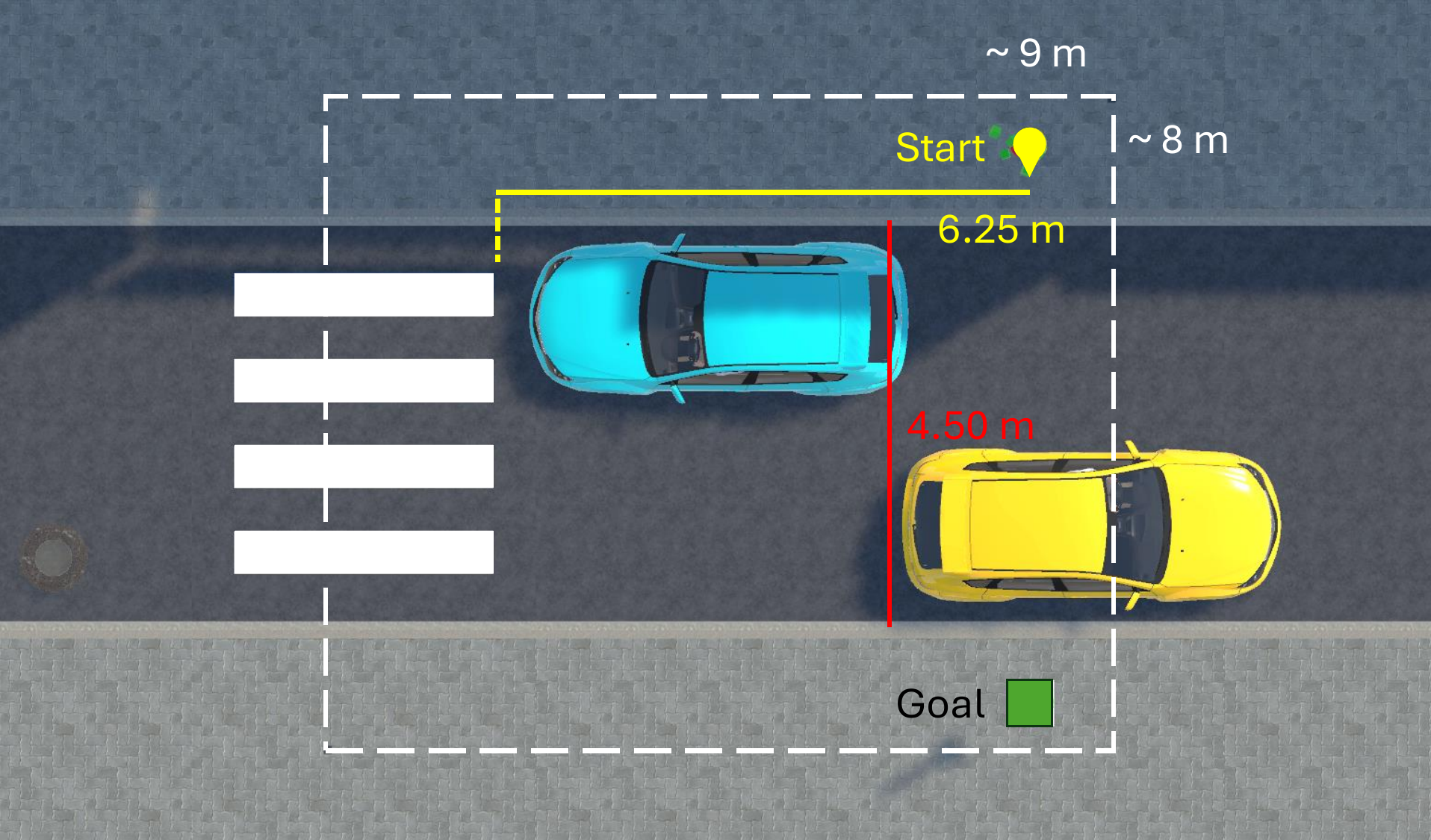}
        \label{fig:schematicZebra}}
    \caption{Schematic overview of the experimental environment. Start (yellow drop) and goal (green square) are visualized and alternated between north and south sides of the road in every other trial. Cars were approaching from both directions with randomly selected gaps per lane. }
    \label{fig:schematicExperiment}
\end{figure*}

We utilize data captured in the virtual reality study by Sprenger et al.~\cite{sprenger2023cross}. This dataset was collected in controlled experiments in Saarbrücken, Germany, and Tokyo, Japan, using a virtual street environment with two lanes and bi-directional traffic. Using an untethered, head-mounted virtual reality headset and motion capture equipment, participants could move freely in a $9\ m\times 8\ m$ space. Within this space, participants could make explicit route choices in different conditions, containing zebra crossings and other virtual pedestrians, and physically navigate the virtual road by walking or running to a designated goal. 
An overview of the experimental setup is illustrated in Fig.~\ref{fig:schematicExperiment}.

Data from 60 participants in each country was collected with equally distributed genders in Sprenger et al.'s study~\cite{sprenger2023cross}, after obtaining approval from the ethical review committee (registration numbers in Japan: H2022-1166-B; and in Germany: 21-08-06). Each participant completed 60 trials, including 15 trials without crossing facilities, 15 trials with a zebra crossing, and 30 trials featuring different combinations of virtual pedestrian avatars crossing on risky and safe gaps and no dedicated crossing facilities. Vehicles in the trials drove at a constant speed of $30\ km/h$, stopping only at the zebra crossing if the participant was nearby. The time gaps between cars were uniformly sampled from 2.5 to 8.5 s.

The experiment recordings allowed for the computation of various parameters describing pedestrian behavior, such as average crossing velocity and selected gap duration, as well as gap-related measures, as listed in Table~\ref{tab:input features}.

\subsection{Prediction Targets and Input Features}
\label{sec:model inputs and outputs}

Using observable information about pedestrian-vehicle interactions before road crossing as inputs, we predict pedestrian behavior at both non-zebra and zebra crossings, as described in Sec.~\ref{sec:data_collection}. In non-zebra crossing scenarios, we investigate pedestrian gap selection behavior, predicting the time gap between cars from both lanes that pedestrians will choose to cross. In zebra crossing scenarios, we analyze pedestrians' decisions regarding the use of zebra crossings and their crossing trajectories.

We mainly use vehicle-gap-related information to predict gap selection and zebra crossing utilization as input features, detailed in Table~\ref{tab:input features} (a).
The computation of gaps in two lanes with opposing traffic is not well defined. For single-lane observations, gaps between vehicles are straightforward to compute by calculating the temporal distance between two vehicles (\textit{car gap near}, \textit{car gap far}). As participants could move laterally to the street, the \textit{effective gap} is influenced by the ego-movement. This \textit{effective gap (near, far)} is measured using an automated virtual stopwatch to record the time between the lead car passing and the following car arriving at the participant's respective spatio-temporal position. 
As the traffic approaches from both directions, the gaps can be additionally computed for both lanes together. Again, this \textit{synchronized gap} can be either computed between cars or as the effective gap, including the ego-movement of the participant. For the \textit{synchronized car gaps}, we always consider the largest gap between vehicles. 
For example, a new gap opens up for the pedestrian, and a vehicle approaches on the near lane from the left, with a distance of 7 s, while another vehicle approaches from the right, with a distance of 5 s. The synchronized gap, including both lanes, is 5 s. If the pedestrian moves to the left, the effective near-lane gap decreases, while the effective far-lane and synchronized gaps increase, giving the participant more time to cross. 
A visualization of these gaps can be found in Fig.~\ref{fig:schematicPedestrians}.

For crossing trajectory prediction, we also consider additional input features when the pedestrian enters the road, as listed in Table~\ref{tab:input features} (b). Unlike existing trajectory prediction algorithms that use past trajectories as input, our study uses only these interaction features without previous position information. This approach allows us to focus on the dynamics of pedestrian-vehicle interactions rather than spatial patterns.

\begin{table}[tb]
    \caption{Features used for modeling, including: (a) Pre-event input variables, all values are measured before entering the road. (b) Additional input variables for trajectory prediction. All values are measured at the frame when entering the road.}
    \label{tab:input features}
    \centering
    \begin{tabular}{m{1.1cm}|m{0.9cm}|m{5.5cm}}
    \hline
    Features & Variable \newline (unit) & Description \\
        \hline
    \multirow{14}{*}{\makecell[c]{(a) \\ Pre-event \\input \\features:\\ measured \\\textit{before} \\entering \\the road.}}    & $T_w (s)$ & Pedestrian waiting time before crossing \\
        & $V_p (m/s)$ & Pedestrian average walking speed \\
        & $N_{en}$ & Number of unused effective gaps at near lane \\
        & $N_{cn}$ & Number of unused car gaps at near lane \\
        & $M_{en} (s)$ & Largest missed effective gap at near lane \\
        & $M_{cn} (s)$ & Largest missed car gap at near lane \\
        & $N_{ef}$ & Number of unused effective gaps at far lane \\
        & $N_{cf}$ & Number of unused car gaps at far lane \\
        & $M_{ef} (s)$ & Largest missed effective gap at far lane \\ 
        & $M_{cf} (s)$ & Largest missed car gap at far lane \\
        & $N_{eb}$ & Number of unused effective gaps for both lanes \\
        & $N_{cb}$ & Number of unused car gaps for both lanes \\
        & $M_{eb} (s)$ & Largest missed effective gap for both lanes \\
        & $M_{cb} (s)$ & Largest missed car gap for both lanes  \\   \hdashline 

    \multirow{5}{*}{\makecell[c]{(b) \\Additional \\input \\features: \\ measured \\\textit{at the} \\\textit{frame} \\\textit{when} \\entering \\the road.}}    & $D_{n} (m)$ & Distance to the vehicle on near lane projected on the lane. \\ 
        & $V_{cn} (m/s)$ & Velocity of the vehicle on near lane (configured as 8.33 m/s, unsmoothed, contains breaking at the zebra crossing). \\ 
        & $D_{f} (m)$ &  Distance to the vehicle on the far lane projected on the lane. \\ 
        & $V_{cf} (m/s)$ & Velocity of the vehicle on far lane (configured as 8.33 m/s, unsmoothed, contains breaking at the zebra crossing).\\ 
        & $D_{z} (m)$ & Distance to the zebra crossing. \\ 
        \hline
    \end{tabular}
\end{table}

\subsection{Predictive Models for Comparing Differences between Countries}
\label{sec:method comparing differences}

Previous studies have indicated significant differences in pedestrian behavior between the data collected in Germany and Japan from a statistical perspective~\cite{sprenger2023cross}. To compare the ground truth pedestrian behavior, we use \texttt{scipy.stats} for statistical tests. 
Non-parametric tests are utilized, as they are more conservative and robust, and not all assumptions for parametric tests (e.g., t-test, ANOVA) are fulfilled. In particular, the Kruskal-Wallis H test (H test) is used instead of a one-way ANOVA to compare differences between different groups, and the Mann-Whitney U test (U test) was used to confirm and further analyze the differences of individual pairs of groups. The acceptance rate was adjusted using Bonferroni correction for multiple comparisons when applicable. 

To investigate the similarities and differences in pedestrian road crossing behavior of data collected in Germany and Japan (RQ1), we conduct a comparative analysis using machine learning models to predict behaviors in each country. We develop separate models for the data collected in Germany and Japan, training and evaluating them independently. Our investigation focuses on comparing the predictability of both models, identifying the most important features used in these models, and understanding the influence of these features on pedestrian behavior.

In comparing differences between countries, we adhere to the methodology outlined by Zhang et al.~\cite{zhang2024predicting}, randomly dividing training and test sets based on trials to ensure no overlaps between training and test sets.
Predicting the accepted gap and crossing trajectories are regression tasks, while predicting zebra crossing usage is a classification task. We utilize the following machine learning models for these regression and classification tasks:

\paragraph{Linear Models}
Linear regression is used for regression as a baseline, considering the linear dependencies between input and output variables. Logistic regression is used for classification as a baseline, predicting the probability of an event by modeling the log-odds for the event as a linear combination of independent input variables.

\paragraph{Support Vector Machine (SVM)}
SVM identifies a hyperplane within the feature space for classification tasks. We use a linear kernel in this work.

\paragraph{Random Forest (RF)}
RF is an ensemble learning method consisting of a large number of regression or decision trees. For regression, it outputs the average predicted results; for classification, it outputs the most selected label. To avoid overfitting, we use 100 estimators with a maximum depth of five.

\paragraph{Neural Networks (NNs)}
NNs, based on a collection of artificial neurons, are used for both regression and classification. Given the small number of input features, we use fully connected neural networks with an input layer, an output layer, and two hidden layers. After using cross-validation to determine the hyper-parameter, we use hidden layer sizes of (2, 4) for gap selection regression networks, (8, 4) for zebra crossing usage classification networks, and (8, 32) for trajectory prediction networks.

\subsection{Model Transferability}
We explore the transferability of predictive models trained on datasets collected in each country (RQ2). To evaluate model transferability, we train models on data collected from one country and evaluate their performance on data from the other country.
To compare the transferability of models, we use two methods for dividing the training and test sets:

\paragraph{Randomly division by trials} Following Zhang et al.'s method~\cite{zhang2024predicting}, we randomly divide the data by trials, ensuring no \textit{trial} overlap between training and test sets.

\paragraph{Randomly division by participant ID} We randomly divide the data by participant ID, ensuring no \textit{participant} overlap between training and test sets. This approach makes prediction more challenging due to the lack of prior participant information compared to the division by trials.

\subsection{Using Cluster Information to Improve Model Transferability}
\label{sec:Using Cluster information to build Transferable Model}

After exploring the differences between countries as described in Sec.~\ref{sec:method comparing differences}, we observed that models trained on datasets from different countries demonstrate different performances, as detailed in Sec.~\ref{sec: gap selection comparison},~\ref{sec: zebra usage comparison}, and~\ref{sec: trajectory comparison}. These differences within the training sets can decrease model transferability across datasets. \textit{Transfer learning}~\cite{torrey2010transfer} is a technique that enhances transferability by enabling models to transfer knowledge from one or several source tasks to new, related target tasks. One possible approach is to increase the similarity between modeling tasks. To address this, we utilize clustering methods, a type of unsupervised learning, to cluster the training data and reduce intra-dataset differences. We explore their potential in improving model performance and transferability (RQ3).

To build models that perform well on datasets collected from both countries, we use unsupervised learning to cluster the input data and consider the cluster information during training. For gap selection and zebra crossing usage, we directly cluster based on the input features. For crossing trajectory prediction, we first cluster the trajectories into groups in the training set and then use this cluster information to train a classifier and a trajectory prediction model. For the test data, we first classify the cluster and then predict the trajectories. The Agglomerative Clustering~\cite{day1984efficient} is used in this work.
This clustering method is compared with three other strategies:

\paragraph{Separate Models} We build two separate models, one for each country. For test data collected from Germany, we use the model trained on data from Germany, and for test data collected from Japan, we use the model trained on data from Japan.

\paragraph{Joint Training} We merge the data collected from both countries and build a single model using the same features as in the separate models. This model is used for predicting on data from both countries.

\paragraph{Country Information as a Feature} We merge the data and build a single model, incorporating country information as a feature in addition to the original input features.

When we have data from both Germany and Japan, we aim to develop a model that performs well in both countries. Considering the practical situation where the same pedestrians are not encountered again, in this section, we divide the training and test sets by participant ID. This ensures the model does not contain any prior information about the same participants in the test set, not even other trials of the same pedestrian.

\subsection{Evaluation Metrics}

To reduce the sensitivity to outliers, we evaluate the performance of gap selection regression models using the mean absolute error (MAE) as defined in Eq.~\ref{eq_mae}, following the evaluation metric employed in Zhang et al.'s study~\cite{zhang2024predicting}. To evaluate the performance relative to the scale of the target variables, we employ the mean absolute percentage error (MAPE) as defined in~\ref{eq_mape}. The ground truth for the $i^{th}$ trial is denoted by $y_i$, the corresponding prediction is denoted by $\hat y_i$, the number of trials is denoted by $n$, and the average value of $y$ over $n$ trials is denoted by $\bar y$.

\begin{equation}
MAE =\frac{\sum_{i=1}^n|\hat y_i - y_i|}{n}
\label{eq_mae}
\end{equation}

\begin{equation}
MAPE(\%) = \frac{MAE}{\bar y} \times 100
\label{eq_mape}
\end{equation}

We evaluate the zebra crossing usage classification models with prediction accuracy (ACC) and F1 score as defined in Eqs.~\ref{eq_acc} and~\ref{eq_f1}, where TP, TN, FP, and FN are the numbers of true positives, true negatives, false positives, and false negatives, respectively. These two metrics are widely employed for evaluating the classification of pedestrian intention, as used in studies~\cite{Fang2018, zhang2021pedestrian, zhang2023cross, zhang2024predicting,Rasouli2019PIE}.

\begin{equation}
ACC=\frac{TP+TN}{TP+FP+TN+FN}
\label{eq_acc}
\end{equation}

\begin{equation}
F1 =\frac{2TP}{2TP+FP+FN}
\label{eq_f1}
\end{equation}

To evaluate trajectory prediction error, we use the Average Displacement Error (ADE), which is widely employed for evaluating pedestrian trajectory prediction, as used in studies~\cite{alahi2016social,gupta2018social,mohamed2020social,zhang2021social,zhang2022learning,zhang2023spatial}.
ADE is defined as the average distance between the ground truth and predicted trajectories over all sampled points. It is given by the following equation, where $Y_j^i$ is the prediction at the $j^{th}$ sampled point in the $i^{th}$ trial, $n$ is the total number of trials, and $m$ is the total number of sampled points for each trial:

\begin{equation}
ADE=\frac{\sum_{i=1}^n\sum_{j=1}^{m}{\| Y_j^i - \hat Y_j^i \|}_2}{n \times m}
\label{eq_ade}
\end{equation}

\subsection{Implementation Details}

After data cleaning, there are a total of 3547 trials of data from Germany, including 880 trials of direct crossing alone, 885 in the risky group, 891 in the safe group, and 891 involving zebra crossings. For data from Japan, there are 3398 trials, with 851 trials of direct crossing alone, 818 in the risky group, 829 in the safe group, and 900 involving zebra crossings.

To better evaluate the models, we use five-fold cross-validation. The dataset is randomly split into five sets. Each time, we use one set for testing and the remaining sets for training, ensuring there is no overlap between training and test data. The reported performance metrics are averaged over the five test sets. Given the differences in scales and types of input variables, we normalize them into the normal distribution to ensure better convergence and stability.

\section{Results and Discussions}

\subsection{Gap Selection Prediction}
\label{sec: gap selection}
\subsubsection{Comparison between Countries}
\label{sec: gap selection comparison}

\begin{table*}[htb]
    \centering
    \caption{The prediction error of gap selection and the three most important features for each model. MAE is in seconds (s), and MAPE is in percentage (\%). A smaller error indicates better performance.}
    \label{tab:regression_direct_alone}
    \begin{tabular}{cc|cc|ccm{1.5cm}m{1.5cm}m{1.5cm}m{1.5cm}}
    \hline
         & & & & \multicolumn{6}{c}{The three most important factors for modeling, out of the 14 pre-event input features} \\ \cline{5-10}
         & Model & MAE (s) & MAPE (\%) & Wait time & Walk speed & \multicolumn{2}{c}{\makecell{Number of unused car/effective \\ (c/e) gaps at far/both (f/b) lanes}} & \multicolumn{2}{c}{\makecell{Largest missed car/effective (c/e) \\ gap at near/both (n/b) lanes}} \\
         & & & & $T_w$ & $V_p$ & \makecell[c]{$N_{cb}$} & \makecell[c]{$N_{ef}$} & \makecell[c]{$M_{cb}$} & \makecell[c]{$M_{en}$} \\ \hline
        \multirow{3}{*}{Germany} & Linear & 1.095 & 18.4 & & \checkmark & \makecell[c]{\checkmark} & \makecell[c]{\checkmark} & & \\
        & RF & 1.091 & 18.4 & \checkmark & \checkmark & & & \makecell[c]{\checkmark} & \\
        & NN & \textbf{1.075} & \textbf{18.1} & \checkmark & & \makecell[c]{\checkmark} & & \makecell[c]{\checkmark} & \\ \hdashline
        \multirow{3}{*}{Japan} & Linear & \textbf{1.030} & \textbf{16.9}  &  & \checkmark & \makecell[c]{\checkmark} & & & \makecell[c]{\checkmark} \\
        & RF & 1.071 & 17.5 & & \checkmark & & & \makecell[c]{\checkmark} & \makecell[c]{\checkmark} \\
        & NN & 1.044 & 17.1 & \checkmark & & \makecell[c]{\checkmark} & & \makecell[c]{\checkmark} & \\
        \hline
    \end{tabular}
\end{table*}

The best-performing models for predicting accepted gaps at non-zebra crossings are the NN model for Germany and the linear regression model for Japan. As shown in Table~\ref{tab:regression_direct_alone}, the NN model trained on German data achieves the best performance with an MAE of 1.075 s and a MAPE of 18.1\%, while the linear regression shows the largest error with an MAE of 1.095 s and a MAPE of 18.4\%. 
In comparison, the linear regression model trained on Japanese data achieved the best results with an MAE of 1.030 s and a MAPE of 16.9\%. This indicates the more non-linear nature of the data from Germany.

\begin{figure}[tb]
    \centering
\includegraphics{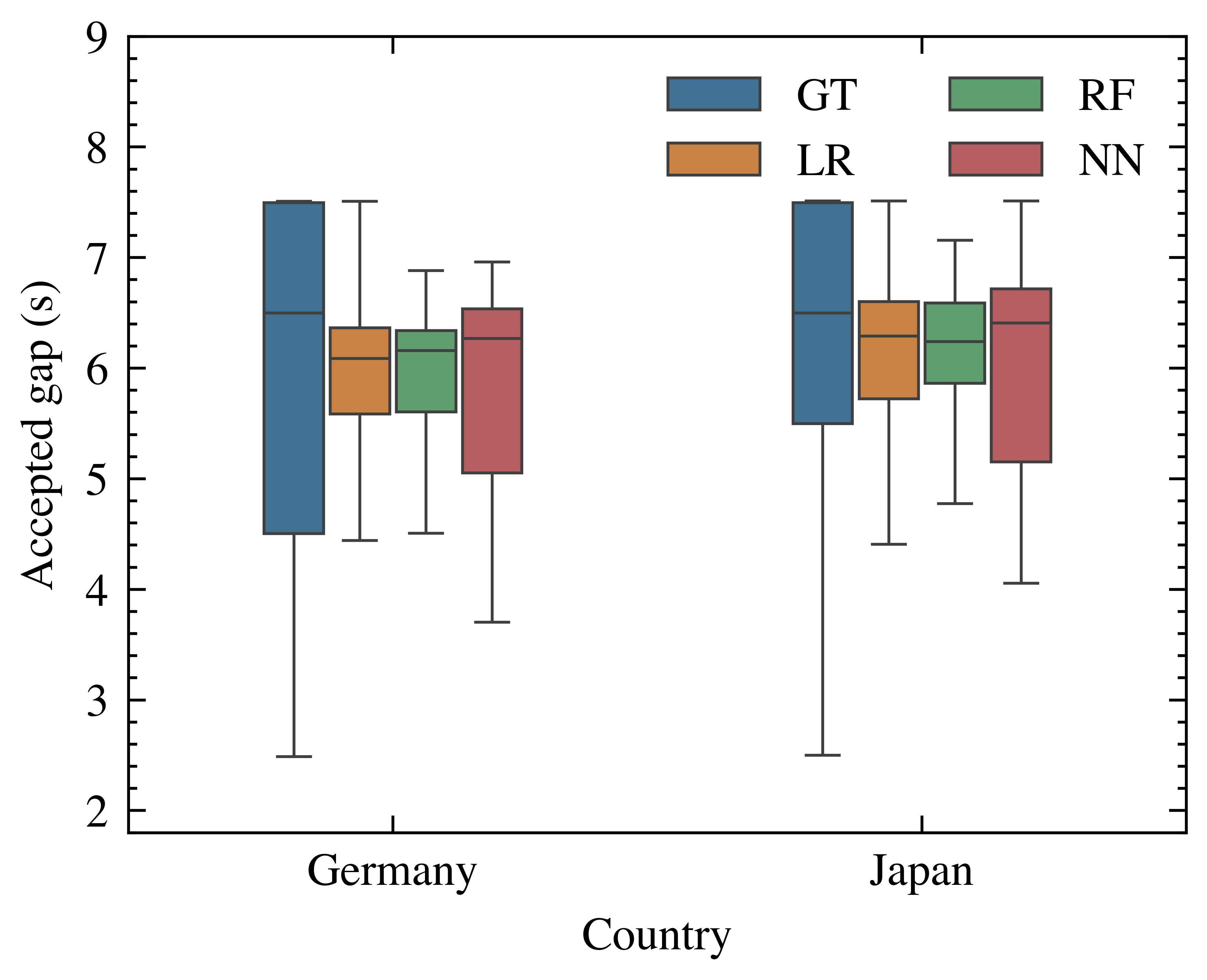}
    \caption{The boxplot of the predicted accepted gaps for different models. GT for ground truth, LR for linear regression, RF for random forest, and NN for neural network.}
    \label{fig:Prediction_selected_gaps_models_overall_distribution_box_compare}
\end{figure}

The NN's predictions are more dispersed and align more closely with the pattern of ground truth, while the prediction results of the linear regression and RF models are concentrated. As demonstrated in Fig.~\ref{fig:Prediction_selected_gaps_models_overall_distribution_box_compare}, the ground truth for the accepted gap in Germany has a median of 6.5 s and an Interquartile Range (IQR) of 3.0 s. The NN's predictions have a median of 6.3 s with an IQR of 1.5 s, while the linear regression's predictions have a median of 6.1 s with an IQR of 0.8 s, and the RF has a median of 6.2 s with an IQR of 0.7 s. This shows that the NN's predictions are closer to the actual ground truth. Similar patterns are observed in the data from Japan.

Data from Germany shows a more dispersed pattern compared to data from Japan. As illustrated in Fig.~\ref{fig:Prediction_selected_gaps_models_overall_distribution_box_compare}, the ground truth for the accepted gap in data from Japan has a median of 6.5 s which is the same with data from Germany. However, the IQR for data from Japan is 2.0 s, which is narrower than the IQR for data from Germany, which is 3.0 s. This indicates greater diversity and complexity in pedestrian behavior in Germany. This may also be the reason for larger prediction errors for the model trained on data collected in Germany. In contrast, data from Japan shows a more concentrated distribution, making it easier to predict, which aligns with the linear regression model's better performance on this dataset.

Different models rely on different key factors to predict pedestrian behavior. As shown in Table~\ref{tab:regression_direct_alone}, NN models consistently consider $N_{cb}$, $T_w$, and $M_{cb}$ as the three most important features across datasets from both countries, indicating NN models' better transferability. In contrast, the key factors differ between models trained on the data from Germany and Japan for the linear and RF models. Specifically, participants in Germany rely more on far lane information, with the linear regression model emphasizing the number of unused effective gaps at the far lane ($N_{ef}$). Meanwhile, participants in Japan focus more on the near lane, as reflected by both the linear regression and RF models emphasizing the largest missed effective gap at the near lane ($M_{en}$).

Despite these differences, the gap selection models in both countries share similar key factors. As shown in Table~\ref{tab:regression_direct_alone}, the number of unused car gaps for both lanes ($N_{cb}$), the largest missed car gap for both lanes ($M_{cb}$), the pedestrian waiting time before crossing ($T_w$), and the pedestrian's average walking speed ($V_p$) consistently emerge as key factors across different models for both countries. Therefore, we analyze the impact of these common features on pedestrian behavior.

\paragraph{Number of unused car gaps}

Significant differences are found in the number of unused car gaps between Germany and Japan (Mann-Whitney U-test, p < .001). The median number of unused car gaps for data collected from Germany is 4.0 with an Interquartile Range (IQR) of 3.0 s, and the median for data from Japan is 4.0 with an IQR of 2.0 s.

\begin{figure}[tb]
\begin{center}
    \subfloat[]{\includegraphics[]{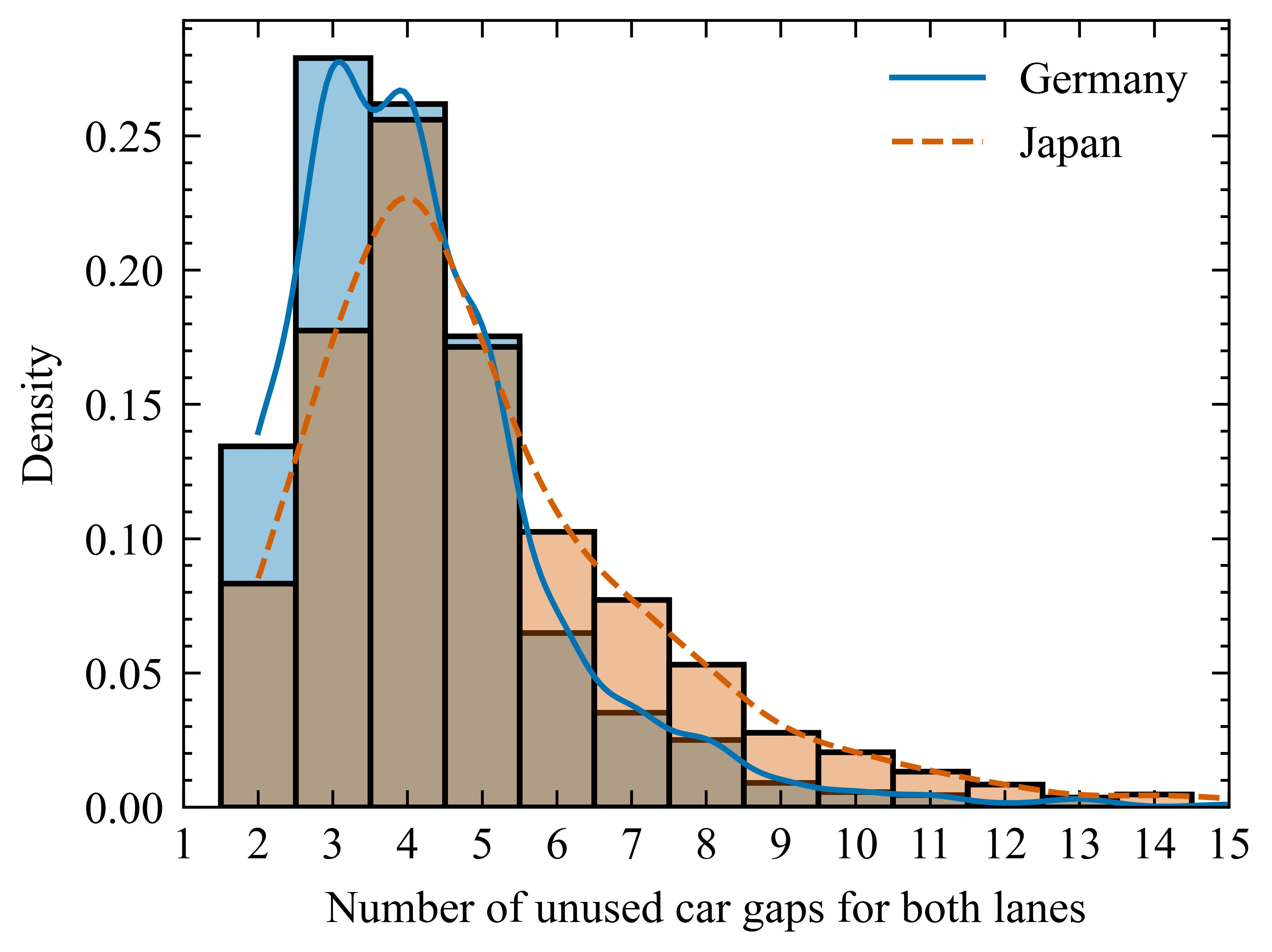}
\label{fig:Unused_sync_cargap_number_distribution}} 
    \hfill
    \subfloat[]{\includegraphics[]{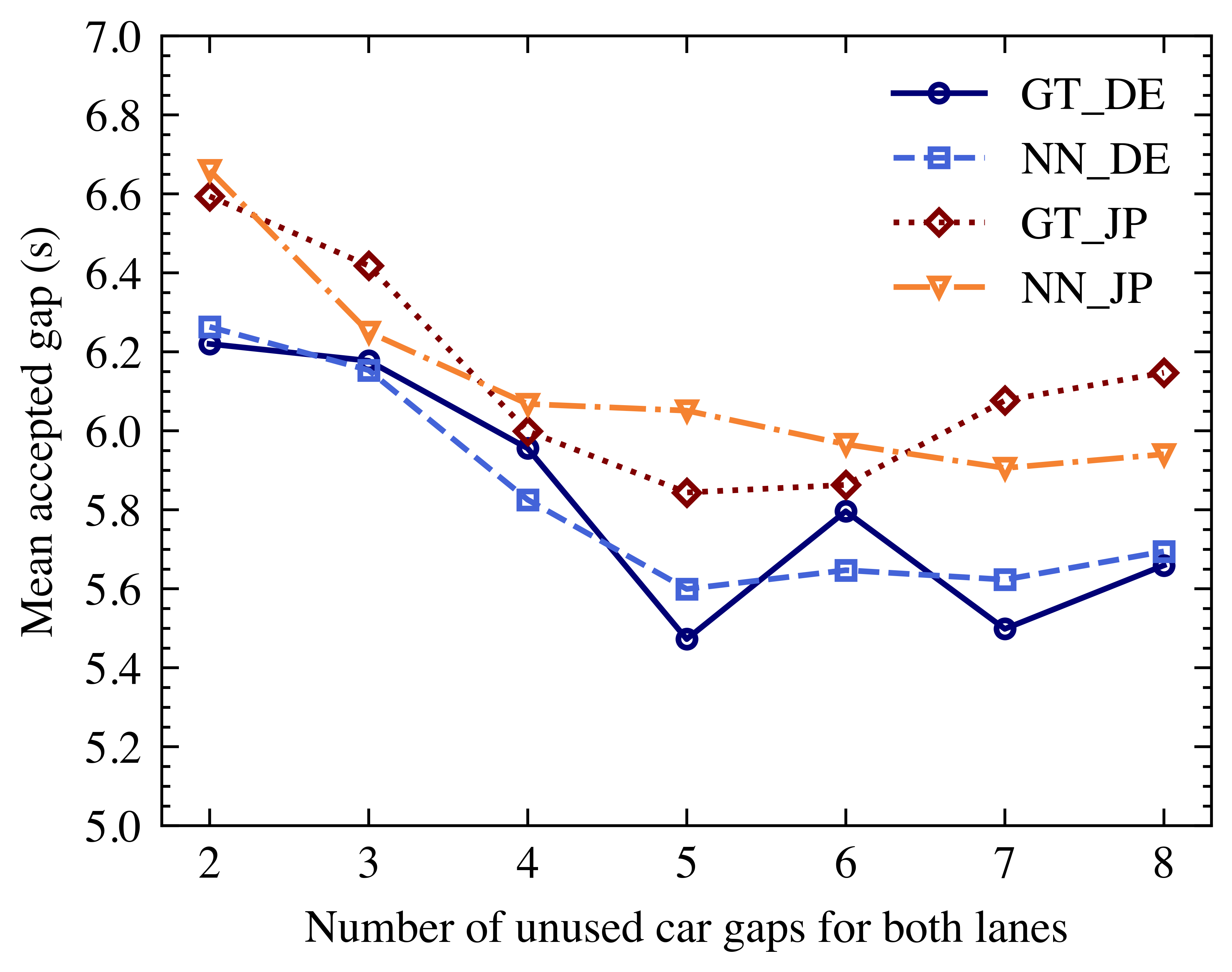}
\label{fig:prediction_accepted_gap_mean_vs_unused_sync_cargap_num}} 
    \caption{(a) Distributions of the number of unused car gaps for both lanes. The pedestrians from the study conducted in Japan missed more car gaps than those from Germany. (b) The mean value of the accepted gap versus the number of unused car gaps for both lanes. DE stands for data collected from Germany, and JP stands for data collected from Japan. For data from both countries, as the number of unused car gaps increases, pedestrians tend to accept smaller gaps.}
    \label{fig:direct_unused_gap}
\end{center}
\end{figure}

Pedestrians in the study conducted in Japan missed more car gaps when crossing compared to those from Germany. Fig.~\ref{fig:Unused_sync_cargap_number_distribution} illustrates the normalized distributions of the number of unused car gaps for both lanes ($N_{cb}$). The distribution peaks at three missed gaps for participants in Germany and four for those in Japan.
When the number of missed car gaps is fewer than or equal to five, participants from Germany show a higher proportion, but when the number exceeds six, participants from Japan show a higher proportion. Overall, this suggests that pedestrians in Japan are more cautious when crossing the road, as they tend to miss more gaps. 

Pedestrians from the study conducted in Japan selected and accepted larger gaps than those in Germany. Fig.~\ref{fig:prediction_accepted_gap_mean_vs_unused_sync_cargap_num} illustrates the relationship between mean accepted gaps and the number of unused car gaps for both lanes. As shown in the figure, the accepted gap for participants in Japan is consistently larger than those in Germany, showing their cautious behavior. 

As the number of unused car gaps increases, the size of accepted gaps tends to decrease for data from both countries. As shown in Fig.~\ref{fig:prediction_accepted_gap_mean_vs_unused_sync_cargap_num}, for example, the mean accepted gap for participants in Germany decreases from 6.2 s to 5.5 s when the number of unused car gaps for both lanes increases from two to five.
This trend suggests that pedestrians are inclined to make riskier choices and accept shorter gaps after missing more gaps.

However, when the number of missed gaps exceeds five, the size of accepted gaps stabilizes. Specifically, for participants in Germany, accepted gaps stabilize between 5.4 and 5.8 s, and for Japan, they stabilize between 5.8 and 6.2 s. This observation implies the existence of a safety threshold for accepted gaps.
Pedestrians tend to cross when the gap is larger than this threshold, as they perceive it to be safe.
Conversely, if a gap falls below this threshold, people may perceive it as unsafe, leading people to avoid crossing.

\begin{figure}[tb]
    \centering
\includegraphics{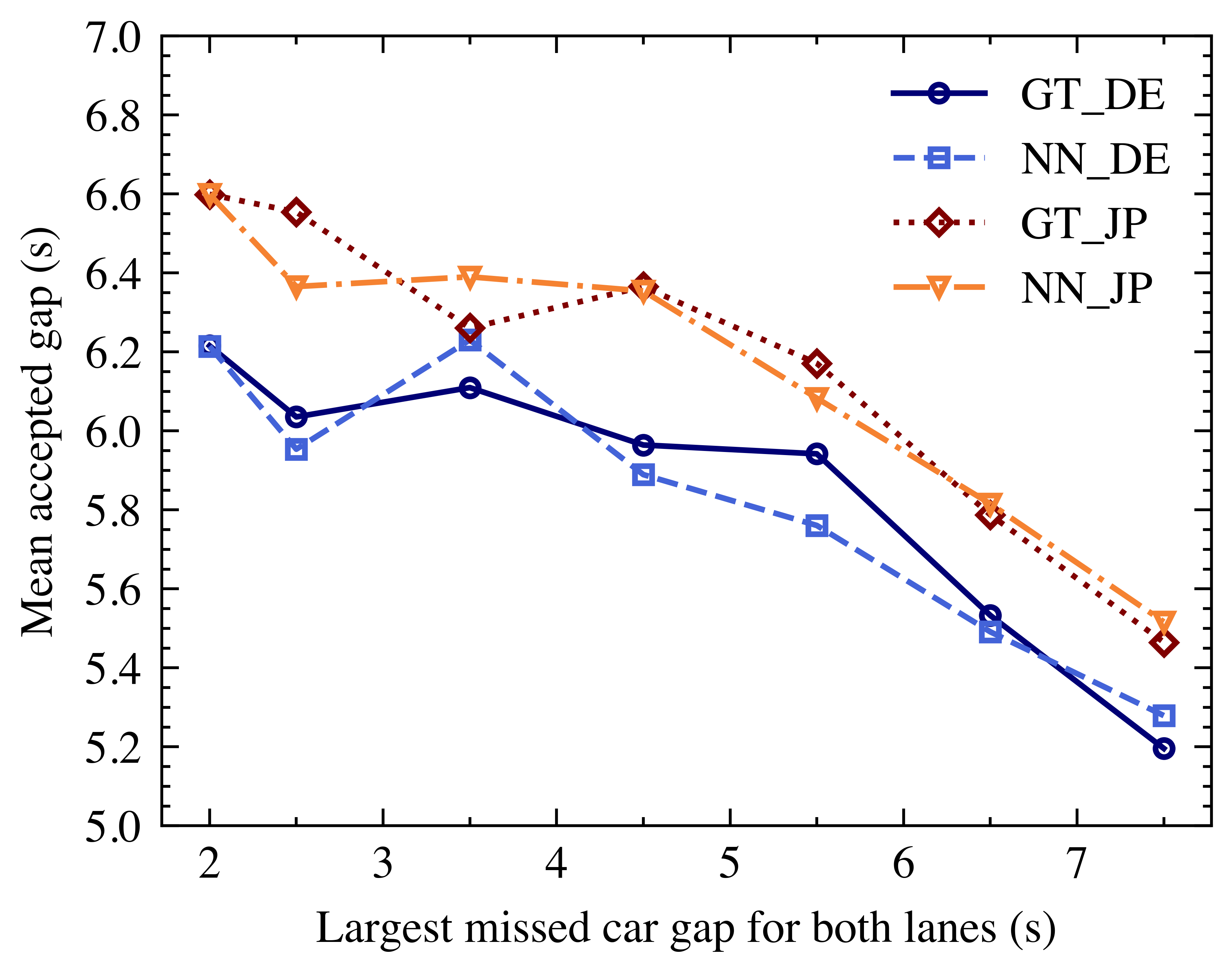}
    \caption{The mean value of the accepted gap versus the largest missed car gap for both lanes. For data from both countries, as the largest missed car gap increases, pedestrians tend to accept smaller gaps.}
    \label{fig:prediction_accepted_gap_mean_vs_largest_missed_sync_gap}
\end{figure}

\paragraph{Largest missed car gap}
When pedestrians miss larger gaps, they tend to choose smaller gaps for crossing. As illustrated in Fig.~\ref{fig:prediction_accepted_gap_mean_vs_largest_missed_sync_gap}, data from both countries show that when the largest missed gap increases, there is an obvious decreasing trend of the accepted gap. This indicates a tendency for pedestrians to make riskier choices after missing crossing chances with larger gaps.
Besides, the larger accepted gaps observed for participants in Japan suggest that they are more cautious when crossing the road compared to those in Germany.

\paragraph{Pedestrian waiting time}

Significant differences are observed in waiting time between the data from Germany and Japan (Mann-Whitney U-test, p < .001). The median waiting time for participants in Germany is 9.52 s (IQR = 5.25 s), whereas for Japan, it is 12.7 s (IQR = 10.11 s).

\begin{figure}[tb]
\begin{center}

    \subfloat[]{\includegraphics[]{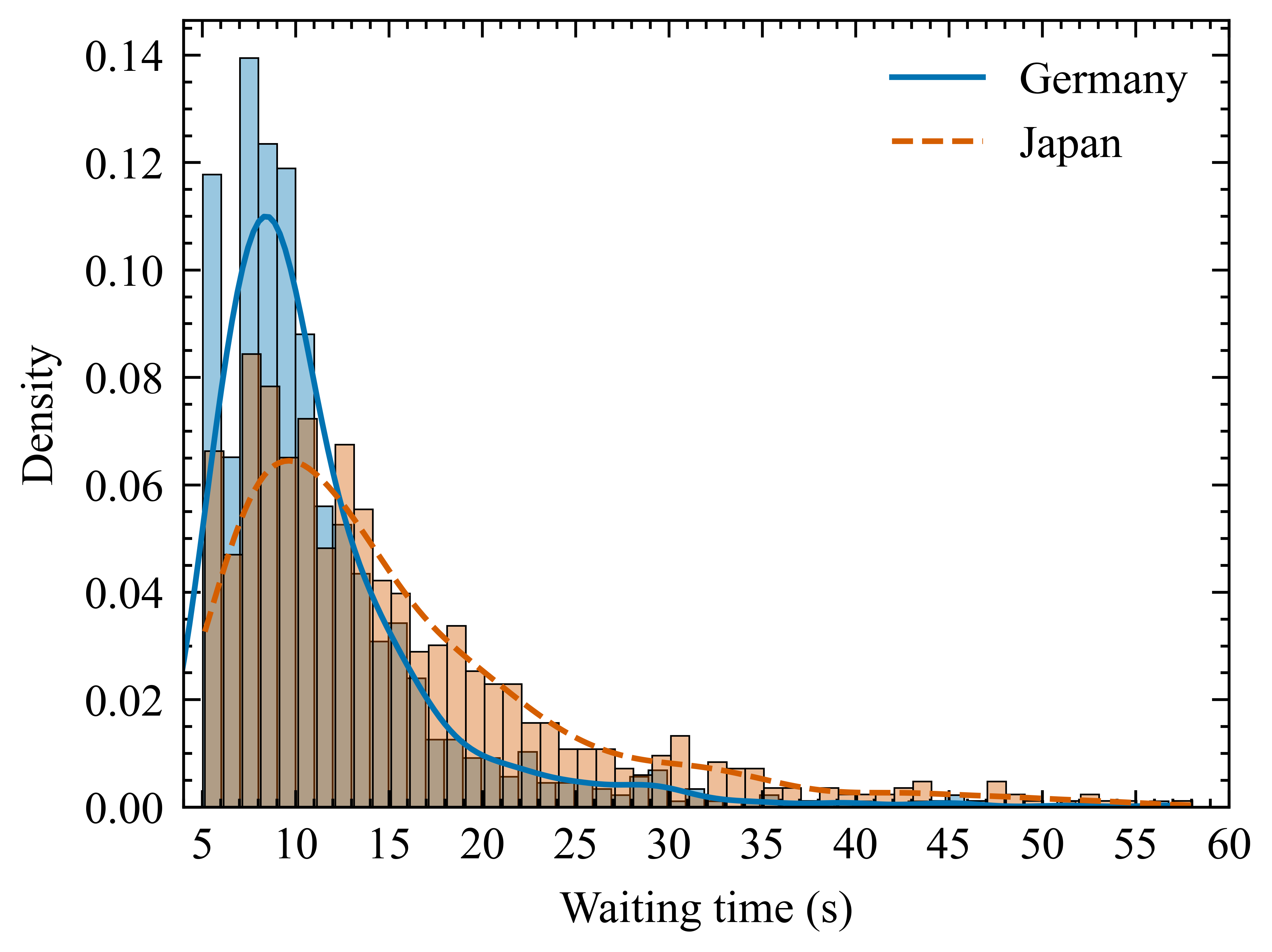}
\label{fig:Waiting_time_distribution_compare_direct_cross}} 
    \hfill
    \subfloat[]{\includegraphics[]{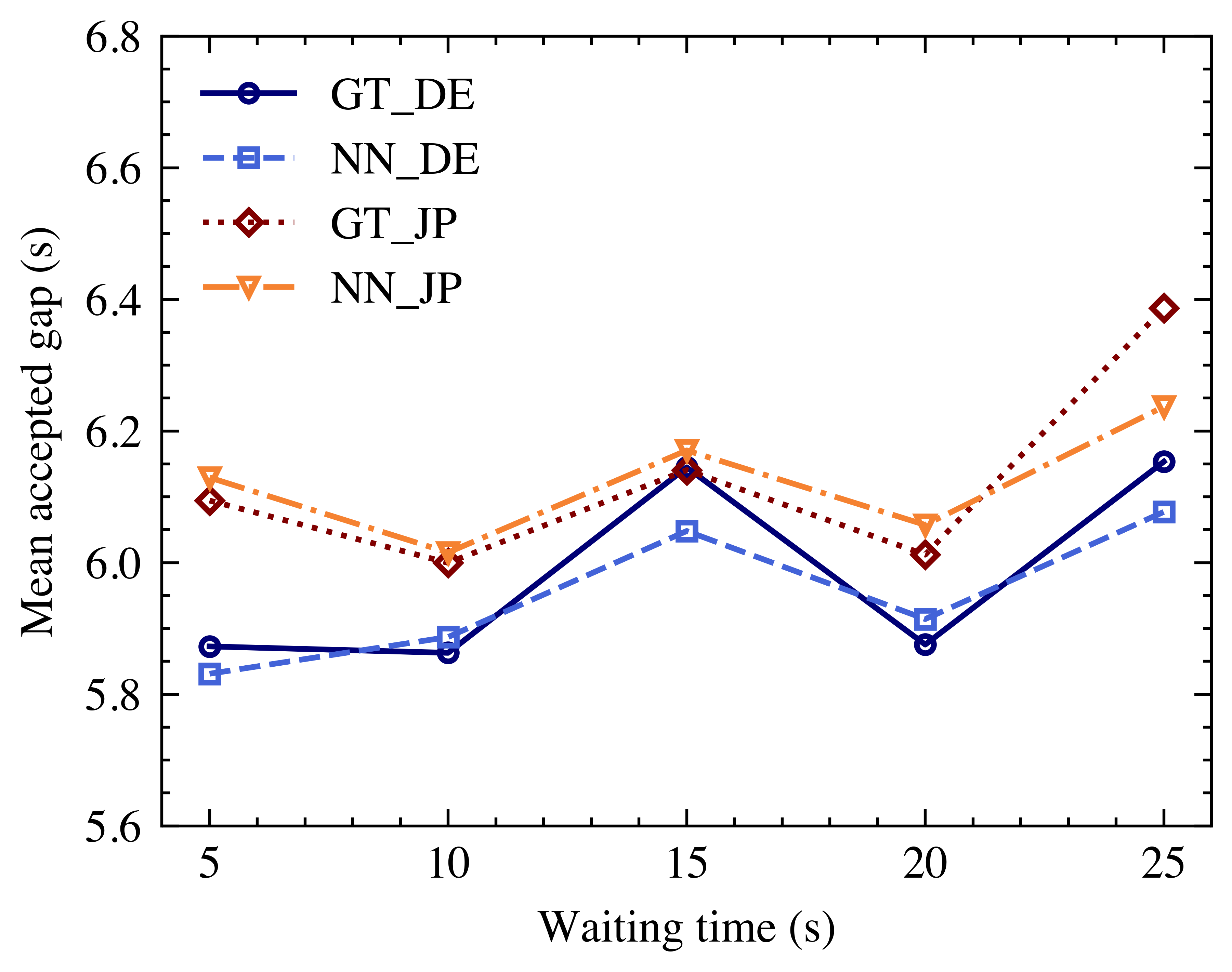}
\label{fig:prediction_accepted_gap_mean_vs_waiting_time}} 
    \caption{(a) Distributions of the pedestrian waiting time. Pedestrians from the data collected in Japan tend to wait longer than participants in the study conducted in Germany. (b) The mean value of the accepted gap versus pedestrian waiting time. For data from both countries, pedestrians tend to select larger gaps when waiting longer.}
    \label{fig:direct_waiting_time}
\end{center}
\end{figure}

Pedestrians in the study conducted in Japan tend to wait longer compared to those in Germany.
Fig.~\ref{fig:Waiting_time_distribution_compare_direct_cross} shows that when the waiting time is shorter than 12 s, participants from Germany show a higher proportion, while when the waiting time exceeds 12 s, participants from Japan show a higher proportion. This pattern indicates a more cautious manner of participants in Japan when crossing the road.

Pedestrians tend to select slightly larger gaps when waiting longer.
Fig.~\ref{fig:prediction_accepted_gap_mean_vs_waiting_time} shows that for data from Germany, the accepted gap increases from 5.9 s to 6.2 s as the waiting time rises from 5 s to 25 s. Similarly, for data from Japan, the accepted gap increases from 6.1 s to 6.4 s over the same waiting time range. This trend contrasts with the tendency to make riskier choices after missing more and larger gaps, indicating that increased waiting time is related to more cautious behavior.
This finding is consistent with the results from Yannis et al.~\cite{yannis2013pedestrian}, which suggests that as pedestrians wait longer to cross the street, the probability of crossing decreases.
Although this might seem counter-intuitive, it can be explained by the fact that pedestrians willing to wait longer are generally more cautious and less likely to take risks. Therefore, those who decide to wait longer tend to choose safer gaps when they eventually cross.

\paragraph{Average walking speed}

\begin{figure}[tb]
\begin{center}

    \subfloat[]{\includegraphics[]{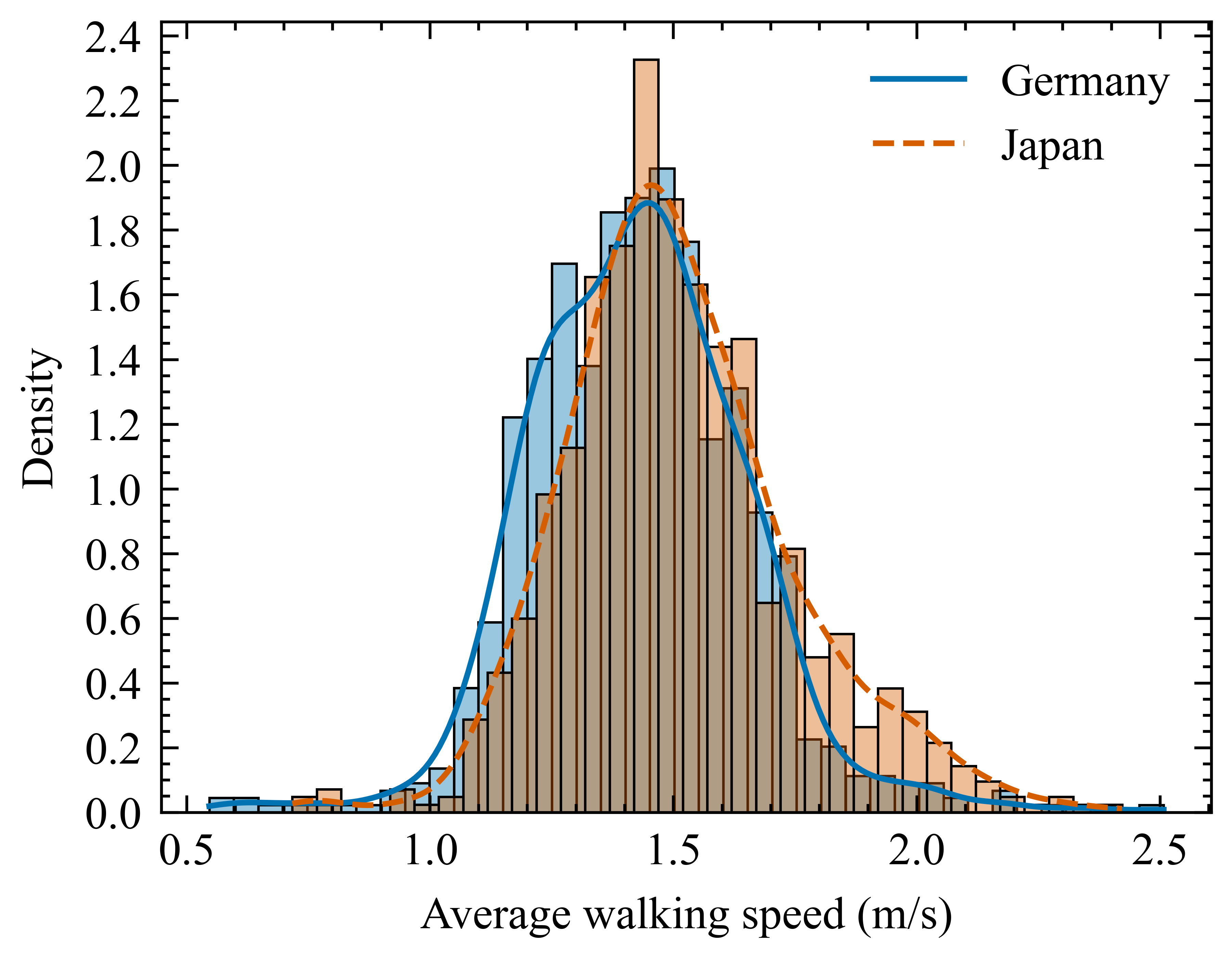}
\label{fig:Walking_speed_distribution_compare}} 
    \hfill
    \subfloat[]{\includegraphics[]{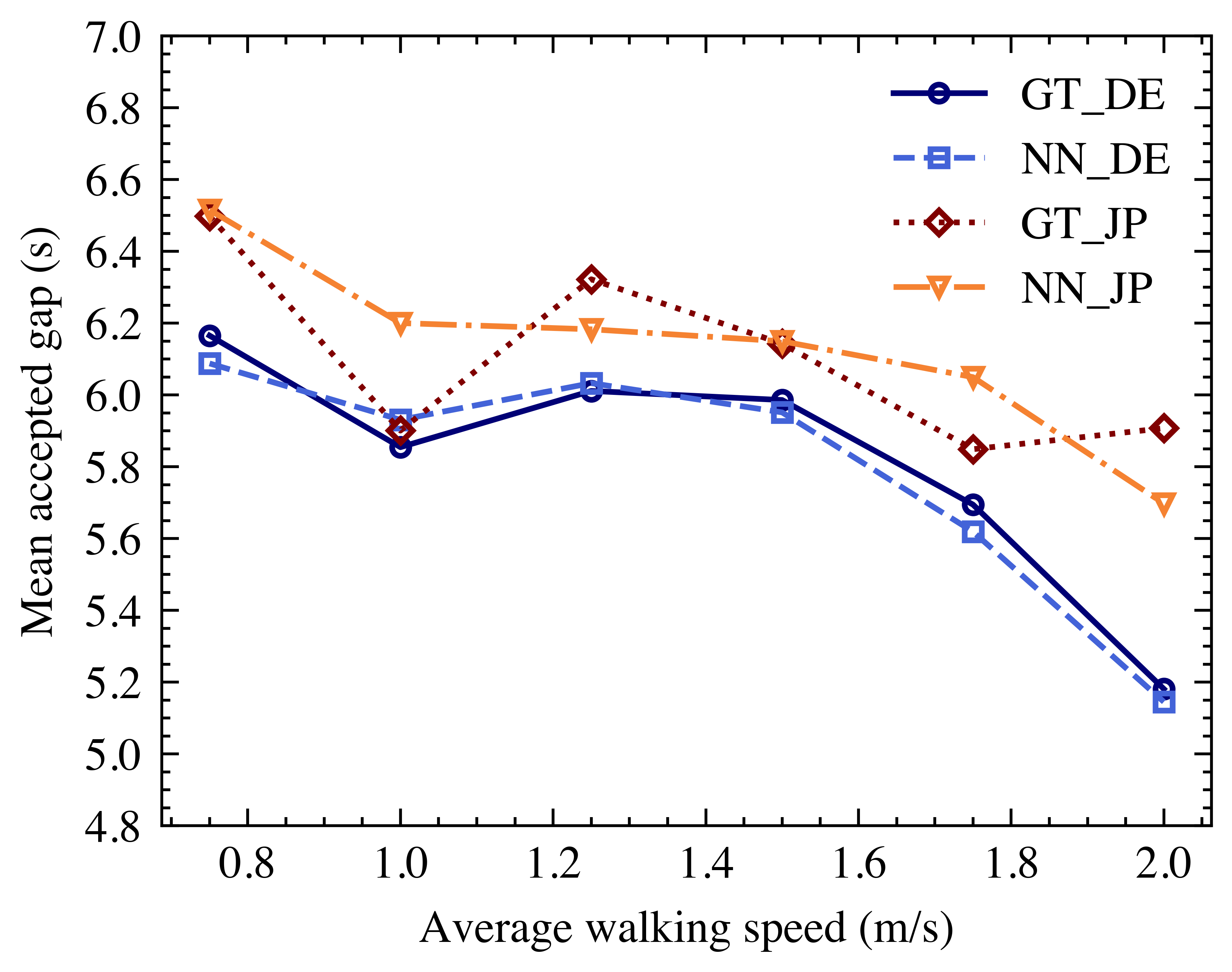}
\label{fig:prediction_accepted_gap_mean_vs_velocity}} 
    \caption{(a) Distributions of the pedestrian walking speed. (b) The mean value of the accepted gap versus pedestrian walking speed. For data from both countries, pedestrians with faster walking speeds tend to choose shorter gaps for crossing.}
    \label{fig:direct_walking_speed}
\end{center}
\end{figure}

There are significant differences in the walking speeds between the data from Germany and Japan (Mann-Whitney U-test, p < .001). 
The median walking speed of participants in Germany is 1.43 m/s (IQR = 0.29 m/s), while in Japan, it is 1.48 m/s (IQR = 0.28 m/s). The statistics and Fig.~\ref{fig:Walking_speed_distribution_compare} indicate that participants in the study conducted in Japan walked slightly faster than those in Germany.

Pedestrians with faster walking speeds tend to choose shorter gaps for crossing. As shown in Fig.~\ref{fig:prediction_accepted_gap_mean_vs_velocity}, participants in Germany who walked at an average speed of 0.75 m/s selected a gap of 6.2 s, whereas those walking at 2 m/s selected a gap of 5.2 s.
This finding is consistent with the study by Wan and Rouphail~\cite{wan2004simulation}, who proposed the critical gap formula for pedestrian crossing behavior, as defined in Eq.~\ref{paper7_eq_cg}:
\begin{equation}
\label{paper7_eq_cg}
    Gap = L/S + F
\end{equation}
where L is the crosswalk length, S is the pedestrian's average walking speed, and F is a safety margin (in seconds) reflecting pedestrian's risk acceptance. The equation shows that as walking speed increases, the critical gap decreases.
An alternative interpretation of this relationship is that individuals in a hurry often prefer shorter gaps, leading to a faster walking speed.

Interestingly, pedestrians from the study conducted in Japan walk slightly faster but still select larger gaps to cross compared to pedestrians from Germany. This behavior could be related to factors such as risk aversion and safety considerations. Although capable of crossing smaller gaps, they choose larger gaps to reduce the risk of accidents or conflicts with vehicles. This preference for larger gaps suggests a safety-conscious behavior.
Another reason for this preference might be to reduce disruptions to traffic flow and show consideration for other road users. By waiting for larger gaps and walking faster, participants from Japan can efficiently cross the road while respecting the movement of vehicles.

\paragraph{Group behavior}

\begin{figure}
    \centering
    \includegraphics{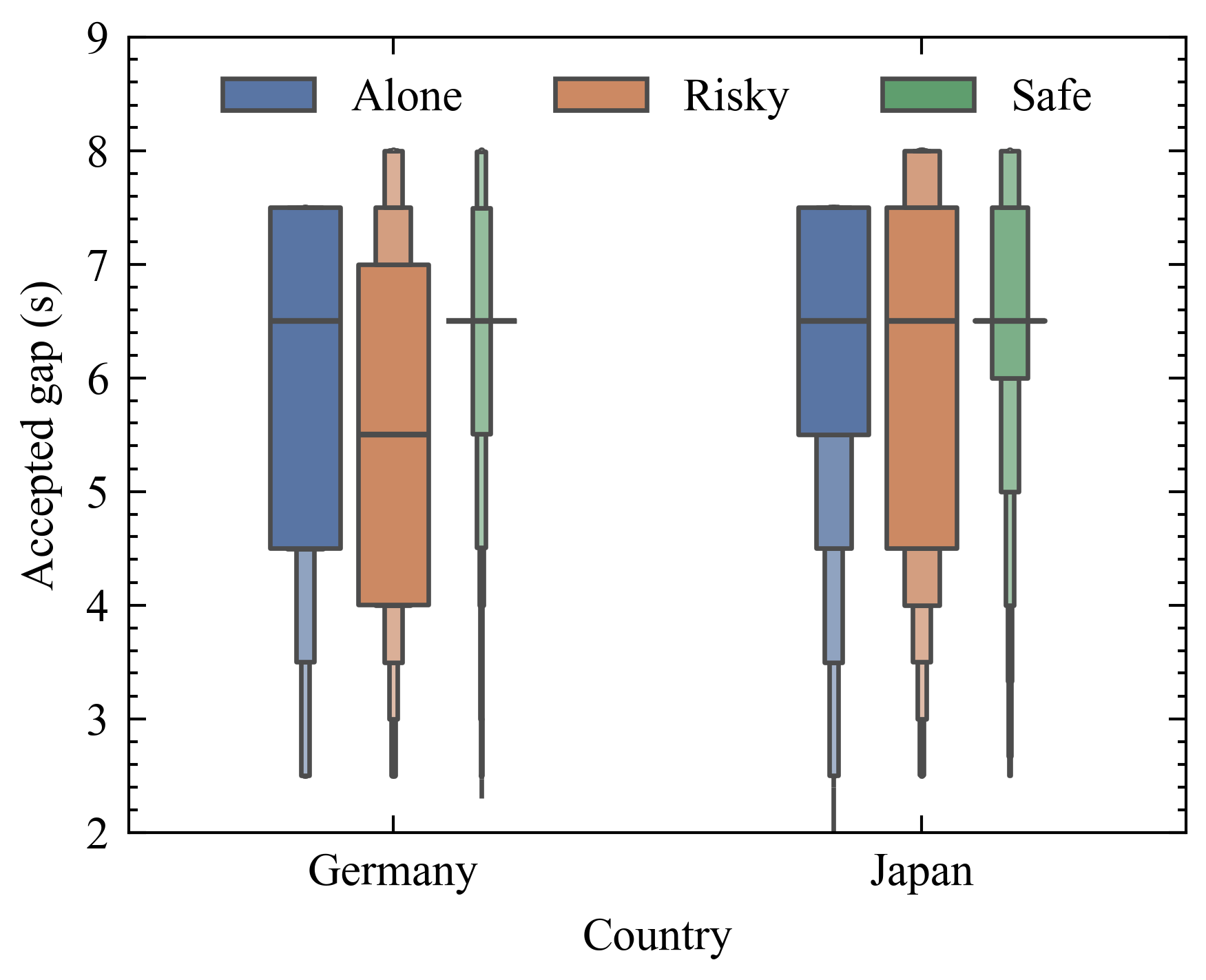}
    \caption{Distribution of the accepted gap for crossing alone, risky group, and safe group by boxen plot (enhanced box plot). It displays the interquartile range (25th to 75th percentile) with additional step boxes showing finer percentiles, including 12.5th to 87.5th percentiles, 6.25th to 93.75th percentiles, and so on. Participants in both countries are influenced by the leading pedestrians, showing a shift towards the gap selected by the leading pedestrians.}
    \label{fig:group_behavior}
\end{figure}

We analyze and compare the distribution of accepted gaps of participants in Germany and Japan under different scenarios (crossing alone, crossing with safe agents, crossing with risky agents).
In the 30 trials of the virtual reality study, groups of one to three virtual pedestrian agents cross alongside the participant, showing two types of behavior: the leading pedestrian agent crosses on a risky gap (4 s gap) or on a safe gap (6.5 s gap).
The accepted participant gap distribution is shown in Fig.~\ref{fig:group_behavior} with an enhanced box plot. The main box displays the interquartile range (25th to 75th percentile), with finer percentiles indicated by additional step boxes.

Both German and Japanese participants show significant differences in the accepted gap among the three groups (German: Kruskal-Wallis H Test, H(2) = 162.7, p < .001; Japanese: Kruskal-Wallis H Test, H(2) = 17.7, p < .001).
Specifically, for participants in Germany, the median accepted gap is 6.5 s (IQR = 3.0 s) for the crossing alone group, 5.5 s (IQR = 3.0 s) for the risky group, and 6.5 s (IQR = 0.0 s) for the safe group. For participants in Japan, the median accepted gap is 6.5 s (IQR = 2.0 s) for the crossing alone group, 6.5 s (IQR = 3.0 s) for the risky group, and 6.5 s (IQR = 0.0 s) for the safe group.

Compared with the scenario of crossing alone, both the safe and risky groups in the study conducted in Germany and Japan indicate a shift towards the gap selected by the leading pedestrian.
For the risky group, participants in the study conducted in Germany exhibit a significantly smaller median accepted gap (5.5 s) compared to crossing alone (6.5 s), with both the 25th and 75th percentiles also lower (4.0 s, 7.0 s for risky group vs. 4.5 s, 7.5 s for crossing alone). For the participants in the study conducted in Japan, the median accepted gap remains consistent across all three groups at 6.5 s. However, the 25th percentile for the risky group is notably smaller (4.5 s) compared to crossing alone (5.5 s).
For the safe group, pedestrians from both countries predominantly followed the leading pedestrians and selected the 6.5 s gap, as indicated by the 25th and 75th percentiles.
This result is consistent with Zhang et al.'s findings~\cite{zhang2024predicting} that pedestrians' gap acceptance is influenced by leading agents.

\paragraph{Other way of dividing the dataset}

These experiments were conducted by dividing the training and test sets randomly by trials. We repeated the experiments by sampling participants into the same age range and dividing the samples randomly by participant ID, obtaining similar results. This consistency indicates that the observed behaviors are influenced by cultural factors rather than age or individual differences.

\subsubsection{Model Transferability}

NN models consistently demonstrated the best transferability. Table~\ref{tab:transferability_direct_cross} shows the results of model transferability evaluated by dividing the dataset by trials and by participant ID. Dividing by participant ID makes prediction more challenging due to the lack of prior knowledge about the pedestrians in the test set, resulting in larger errors compared to dividing by trials.
NN models achieved the highest accuracy when tested on the other country's dataset, regardless of how the dataset was divided. As noted earlier, the top three important features were the same for NN models in both countries. This consistency suggests that NN models rely on the same variables for decision-making, contributing to their better transferability.

Interestingly, the error decreased when the NN model trained on data from Germany was applied to data from Japan. This implies that the behavior patterns of participants in Germany are more varied and harder to predict, while the behavior patterns of participants in Japan are more consistent and easier to predict.

\begin{table}[htb]
\caption{Transferability of models. Upper: divided by trials, lower: divided by participant ID. MAE is in seconds (s), and MAPE is in percentage (\%). A smaller error indicates better performance.}
\label{tab:transferability_direct_cross}
\begin{center}
\begin{tabular}{cccccc}
\hline
\multirow{2}{*}{Divided by} & \multirow{2}{*}{Model} & \multicolumn{2}{c}{Germany to Japan} & \multicolumn{2}{c}{Japan to Germany} \\
 &  & \multicolumn{1}{l}{MAE} & \multicolumn{1}{l}{MAPE} & \multicolumn{1}{l}{MAE} & \multicolumn{1}{l}{MAPE} \\
 \hline
\multirow{3}{*}{\makecell{Trials}} & Linear & 1.080 & 17.7 & 1.099 & 18.5 \\
 & RF & 1.121 & 18.3 & 1.118 & 18.8 \\
 & NN & \textbf{1.054} & \textbf{17.3} & \textbf{1.086} & \textbf{18.3} \\
 \hdashline
\multirow{3}{*}{\makecell{Participant \\ID}} & Linear & 1.082 & 17.7 & 1.102 & 18.6 \\
 & RF & 1.134 & 18.6 & 1.126 & 19.0 \\
 & NN & \textbf{1.069} & \textbf{17.5} & \textbf{1.097} & \textbf{18.5} \\
 
 \hline
\end{tabular}
\end{center}
\end{table}

\subsubsection{Using Clustering to Improve Performance and Transferability}

Using clustering information helps to capture shared patterns of pedestrian behavior across different countries, resulting in improved model performance.
We compare our proposed model, which uses clustering information to separate models, joint training models, and the model using country information as a feature. The dataset is divided into training and test sets randomly by participant ID.
As shown in Table~\ref{tab:model_results_transferability}, when mixing the data from both countries, the overall error decreases compared with separate models. Adding country information slightly improves the results compared to separate models, but not as much as joint training. Incorporating cluster information yields the best average results with an error of 17.5\%.

\begin{table}[htb]
\centering
\caption{Results for transferable models for gap selection behavior prediction. NN models are used. MAE is in seconds (s), and MAPE is in percentage (\%). A smaller error indicates better performance.}
\label{tab:model_results_transferability}
\begin{tabular}{ccccccc}
\hline
\multirow{2}{*}{Strategy} & \multicolumn{2}{c}{German Test} & \multicolumn{2}{c}{Japanese Test} & \multicolumn{2}{c}{Average} \\
 & \multicolumn{1}{l}{MAE} & \multicolumn{1}{l}{MAPE} & \multicolumn{1}{l}{MAE} & \multicolumn{1}{l}{MAPE} & \multicolumn{1}{l}{MAE} & \multicolumn{1}{l}{MAPE} \\
 \hline
Separate & 1.132 & 19.1 & \textbf{1.032} & \textbf{16.9} & 1.082 & 18.0 \\
Joint & 1.080 & 18.2 & 1.049 & 17.2 & 1.064 & 17.7 \\
Country & 1.103 & 18.6 & 1.049 & 17.2 & 1.076 & 17.9 \\
Cluster & \textbf{1.067} & \textbf{18.0} & 1.043 & 17.1 & \textbf{1.055} & \textbf{17.5} \\
\hline
\end{tabular}
\end{table}

\subsection{Zebra Crossing Usage Prediction}
\subsubsection{Comparison between Countries}
\label{sec: zebra usage comparison}
NNs outperform other models in predicting whether pedestrians use the zebra crossings.
The accuracy and F1 score of the prediction results are presented in Table~\ref{tab:accuracy_zebra_usage_both_country}. NNs outperform logistic regression, SVM, and RF for data from both countries, achieving an accuracy rate of 94.27\% in data from Germany and 94.26\% in data from Japan, showing their effectiveness in capturing behavior patterns.

\begin{table*}[htb]
    \centering
    \caption{The prediction accuracy and F1 score for the use of zebra and the three most important features for each model. Accuracy and F1 in percentage (\%). A higher score indicates better performance.}
    \label{tab:accuracy_zebra_usage_both_country}
    \begin{tabular}{cc|cc|ccm{1.5cm}m{1.5cm}m{0.7cm}m{0.7cm}m{0.7cm}m{0.7cm}}
    \hline
         & & & & \multicolumn{8}{c}{The three most important factors for modeling, out of the 14 pre-event input features} \\ \cline{5-12}
         & Model & ACC (\%) & F1 score (\%)& \makecell{Wait time} & \makecell{Walk speed} & \multicolumn{2}{c}{\makecell{Number of unused car/effective \\ (c/e) gaps at near (n) lanes}} & \multicolumn{4}{c}{\makecell{Largest missed car/effective \\ (c/e) gap at near/far (n/f) lanes}} \\
         & & & & $T_w$ & $V_p$ & \makecell[c]{$N_{cn}$} & \makecell[c]{$N_{en}$} & \makecell[c]{$M_{cn}$} & \makecell[c]{$M_{en}$} & \makecell[c]{$M_{ef}$} & \makecell[c]{$M_{eb}$} \\ 
         \hline
         
        \multirow{4}{*}{Germany} & Logistic & 90.25 & 89.22 & & \makecell[c]{\checkmark} & & \makecell[c]{\checkmark} & \makecell[c]{\checkmark} & & & \\
        & SVM & 91.74 & 91.08 & & & \makecell[c]{\checkmark} & \makecell[c]{\checkmark} & \makecell[c]{\checkmark} & & & \\
        & RF & 91.85 & 91.73 & \makecell[c]{\checkmark} & & & & & & \makecell[c]{\checkmark} & \makecell[c]{\checkmark} \\
        & NN & \textbf{94.27} & \textbf{93.91} & \makecell[c]{\checkmark} & & & & \makecell[c]{\checkmark} & \makecell[c]{\checkmark} & &  \\
        \hdashline
        
        \multirow{4}{*}{Japan} & Logistic  & 90.66 & 90.96 & & & & \makecell[c]{\checkmark} & \makecell[c]{\checkmark} & \makecell[c]{\checkmark} & &\\
        & SVM  & 92.46 & 92.79 & & & \makecell[c]{\checkmark} & \makecell[c]{\checkmark} & \makecell[c]{\checkmark} & & & \\
        & RF & 91.55 & 92.19 & \makecell[c]{\checkmark} & & & & & \makecell[c]{\checkmark} & \makecell[c]{\checkmark} \\
        & NN & \textbf{94.26} & \textbf{94.64} & \makecell[c]{\checkmark} & & & \makecell[c]{\checkmark} & \makecell[c]{\checkmark} & & & \\
        \hline
    \end{tabular}
\end{table*}

Linear and non-linear-based models rely on different key features as shown in Table~\ref{tab:accuracy_zebra_usage_both_country}. Specifically, linear-based models, including logistic regression and SVM, identify the number of unused effective gaps in the near lane ($N_{en}$) as the most important feature for data from both countries.
In contrast, non-linear models, including RF and NN, prioritize pedestrian waiting time ($T_w$) as the most critical feature for modeling data from both countries. Therefore, we investigate the impact of these two factors ($N_{en}$ and $T_w$) on prediction accuracy.

\paragraph{Number of unused effective gaps}

\begin{figure}
    \centering
    \includegraphics{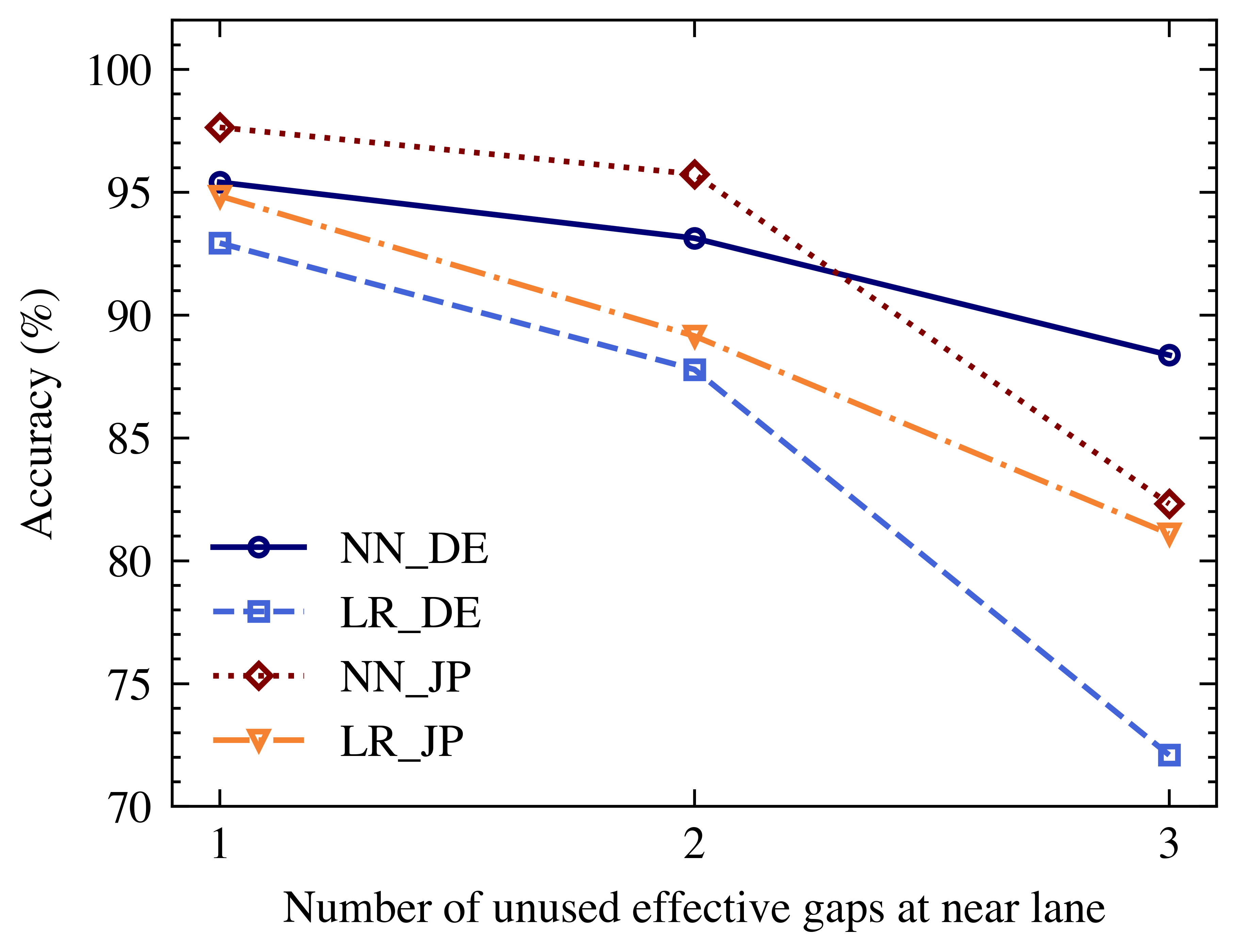}
    \caption{The prediction accuracy versus the number of unused effective gaps at the near lane. NN stands for the results of neural networks, and LR stands for the results of Logistic regression. As the number of unused effective gaps at the near lane increases, prediction accuracy decreases for data from both countries.}
    \label{fig:prediction_accuracy_vs_eff_near_gap_num}
\end{figure}

As the number of unused effective gaps at the near lane increases, prediction accuracy decreases for data from both countries.
Fig.~\ref{fig:prediction_accuracy_vs_eff_near_gap_num} shows that for data in Germany, the accuracy of the NN model drops from 95.4\% to 88.4\% as the number of unused effective gaps at the near lane rises from one to three. Similarly, logistic regression accuracy decreases from 92.9\% to 72.1\%. 
Similar trends are observed for data in Japan.
This indicates the increased difficulty in predicting zebra crossing usage when the number of unused gaps increases. Notably, neural networks consistently demonstrate high accuracy across all gap numbers compared to other models, demonstrating their capability of prediction.

\paragraph{Pedestrian waiting time}

\begin{figure}[tb]
\begin{center}

    \subfloat[]{\includegraphics[]{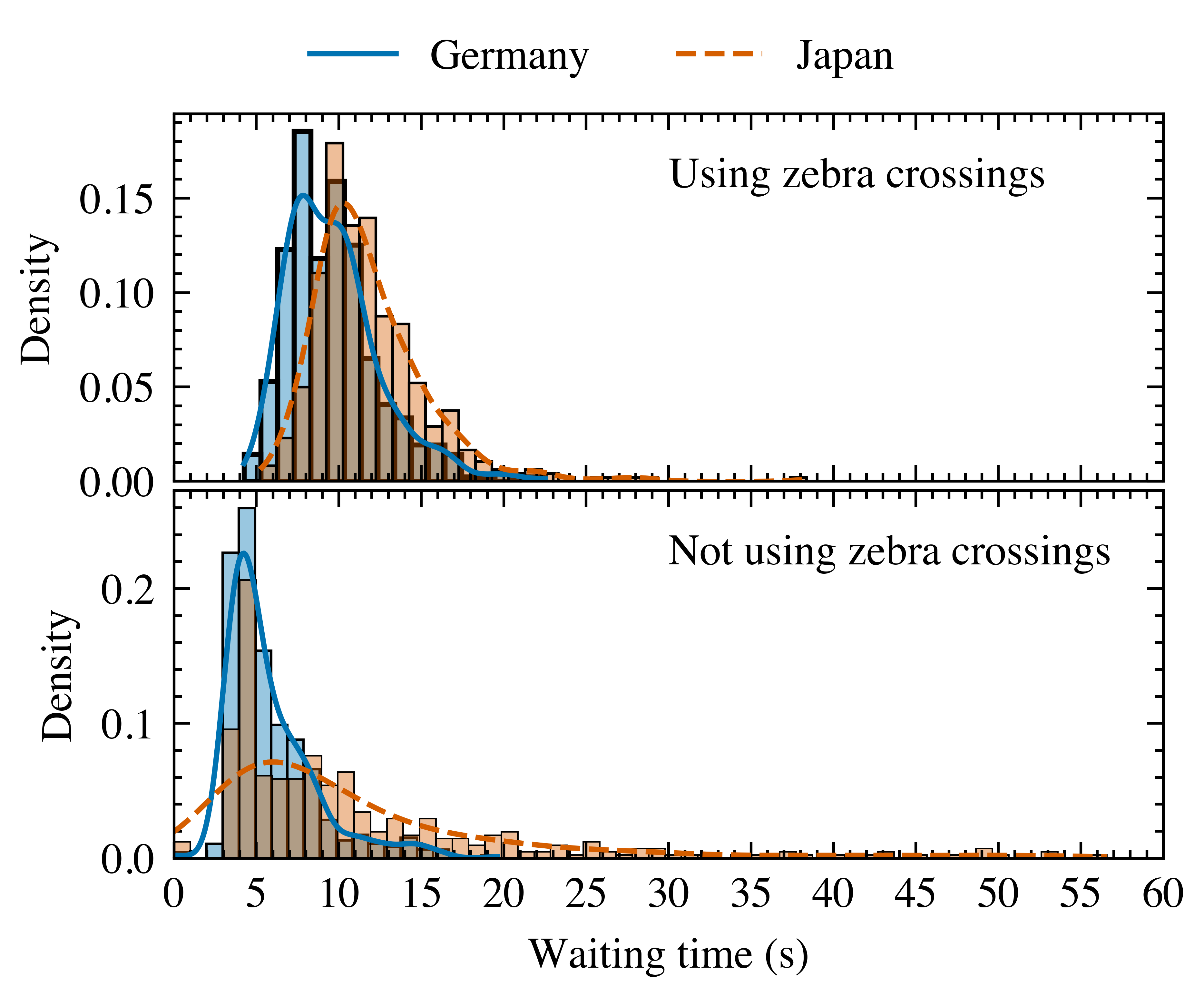}
\label{fig:Zebra_and_non_zebra_waiting_time_distribution_compare}} 
    \hfill
    \subfloat[]{\includegraphics[]{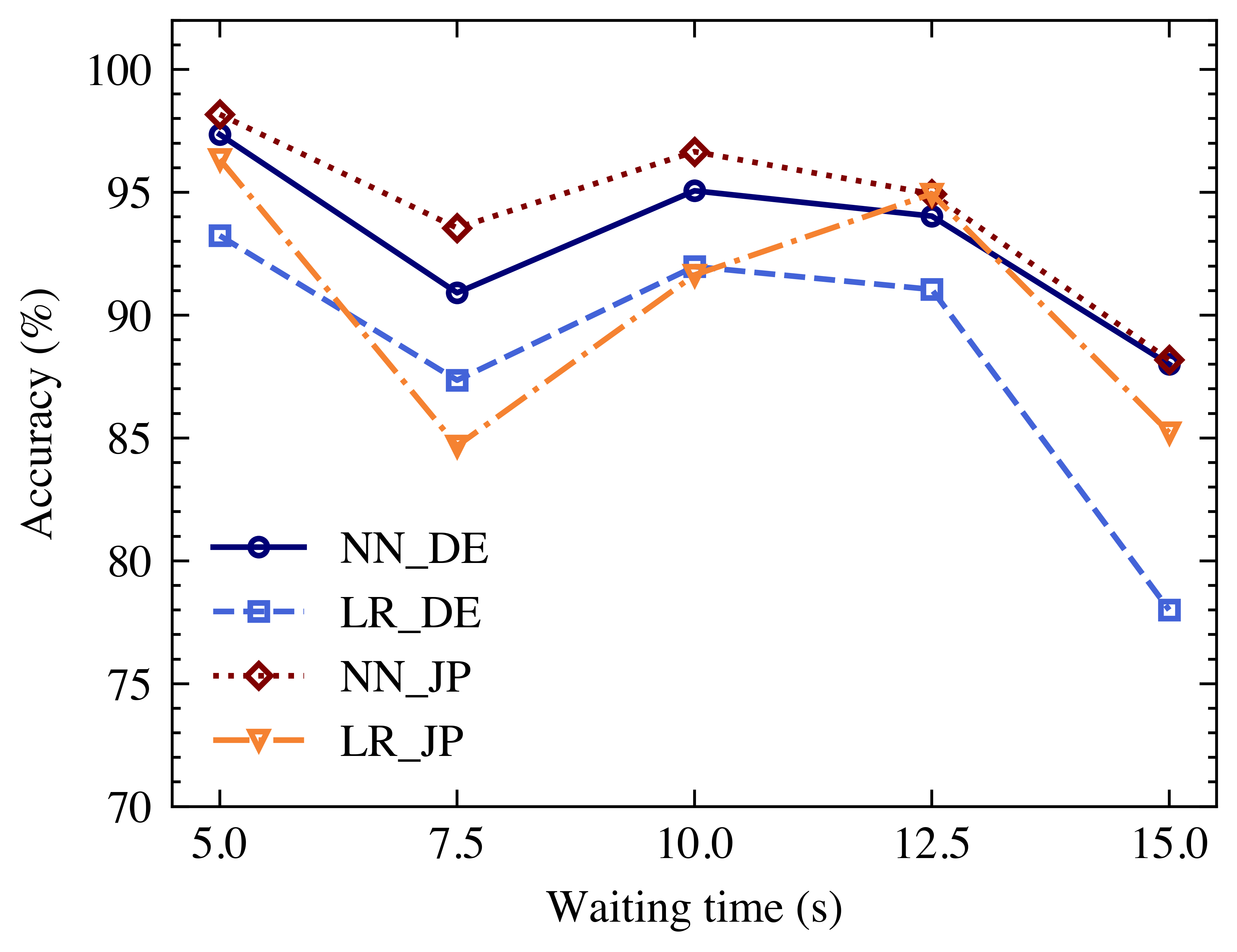}
\label{fig:prediction_accuracy_vs_waiting_time}}
    \caption{(a) The distributions of pedestrian waiting time for zebra crossing cases. Upper: cases where pedestrians used zebra crossing. Lower: cases where pedestrians did not use zebra crossings. (b) The prediction accuracy versus pedestrian waiting time. With increasing waiting time, the predictability of zebra crossing usage decreases for data from both countries.}
    \label{fig:zebra_waiting_time}
\end{center}
\end{figure}

Significant differences are observed in waiting time between the data from Germany and Japan (Mann-Whitney U-test, p < .001).
Overall, pedestrians from data collected in Japan exhibit longer waiting times (median = 10.3 s, IQR = 5.6 s) compared to those from Germany (median = 7.3 s, IQR = 4.9 s), indicating a higher level of caution.
The distributions of pedestrian waiting time for zebra crossing cases are illustrated in Fig.~\ref{fig:Zebra_and_non_zebra_waiting_time_distribution_compare}. The upper section corresponds to cases where pedestrians used zebra crossings, while the lower section represents cases where pedestrians did not use zebra crossings. For both groups, participants in Japan show longer waiting times.

With increasing waiting time, the predictability of zebra crossing usage decreases for both countries. Fig.~\ref{fig:prediction_accuracy_vs_waiting_time} shows that for data in Germany, the prediction accuracy of the NN model decreases from 97.4\% to 88.0\% when the waiting time increases from 5.0 s to 15.0 s. Similarly, the accuracy of the logistic regression model decreases from 93.2\% to 78.0\%. 
This suggests that as pedestrians wait longer, their decisions become more influenced by implicit factors, making them harder to predict.
Neural networks consistently exhibit better results than linear models, showing their capability to handle nonlinearity.

\paragraph{Accepted gaps}

\begin{figure}
    \centering
    \includegraphics{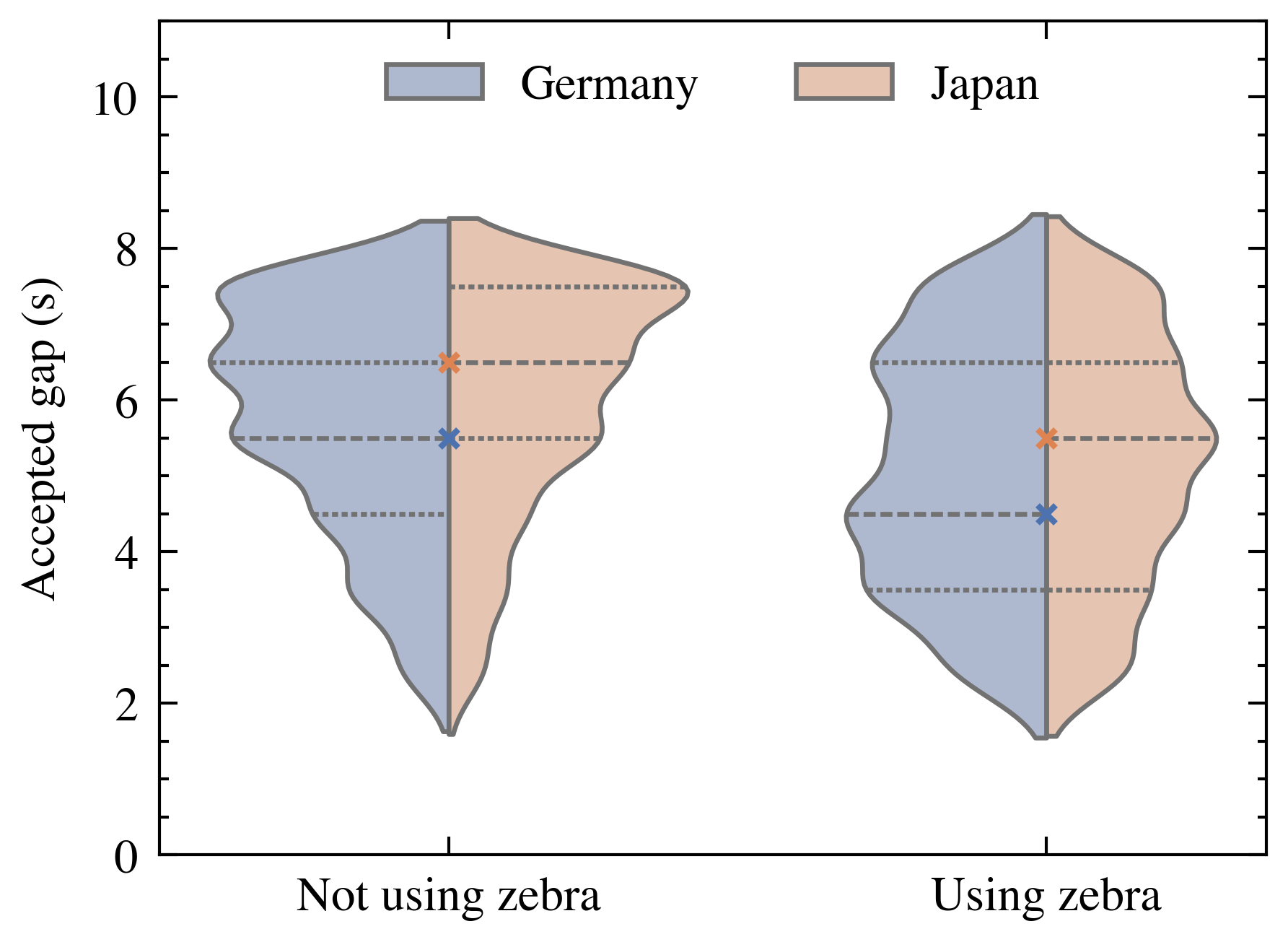}
    \caption{The distributions of the accepted gaps for pedestrians who used and did not use zebra crossings from both countries. Violin plots are shown with quartiles representing the 25th (top line), 50th (middle line), and 75th (bottom line) percentiles of the distribution. The cross marks indicate the median value of the distributions. Pedestrians from the study conducted in Japan accepted larger gaps than those from Germany for both zebra crossing and non-zebra crossing scenarios.
    }
    \label{fig:accepted_gap_vs_zebra_using_vs_country}
\end{figure}

Pedestrians who use the zebra crossing generally accept smaller gaps than those who do not in both countries. Fig.~\ref{fig:accepted_gap_vs_zebra_using_vs_country} compares the distribution of accepted gaps between Germany and Japan, along with pedestrians' zebra crossing usage. As shown in Fig.~\ref{fig:accepted_gap_vs_zebra_using_vs_country}, participants in Germany who used zebra crossings had a median waiting time of 4.5 s, compared to 5.5 s for those who did not use them. In Japan, the median waiting time was 5.5 s for those using zebra crossings and 6.5 s for those not using them.
This suggests pedestrians are inclined not to use the zebra crossing when larger gaps are available. Besides, the presence of a zebra crossing likely reduces pedestrians' risk aversion, expecting vehicles to yield or stop.

Comparing accepted gaps between pedestrians from the studies conducted in Germany and Japan, we find that participants in Germany accept smaller gaps than those in Japan, irrespective of zebra crossing usage. This indicates a more cautious behavior of Japanese participants compared to German participants. 

\subsubsection{Model Transferability}

NN models consistently achieve the best performance and transferability for data from both countries. As shown in Table~\ref{tab:zebra_usage_acc_trials}, NN models outperform other models in predicting zebra crossing usage, regardless of whether the datasets are divided by trials or by participant ID. Specifically, when transferring from Germany to Japan, the NN model achieves an accuracy of 91.67\%, and when transferring from Japan to Germany, it achieves an accuracy of 91.32\% when training and testing sets are divided by participant ID.
When using the NN model, dividing the data by participant ID results in higher accuracy compared to dividing the data by trials. This suggests that dividing by participant ID during training leads to better transferability, as it helps to avoid overfitting and allows the model to generalize better to unseen individuals.

\begin{table}[tb]
\caption{Results of Transferability. In percentage (\%). A higher score indicates better performance.}
\label{tab:zebra_usage_acc_trials}
\begin{center}
\begin{tabular}{llcccc}
\hline
\multirow{2}{*}{Divided by} & \multirow{2}{*}{Model} & \multicolumn{2}{c}{Germany to Japan} & \multicolumn{2}{c}{Japan to Germany} \\
 &  & \multicolumn{1}{c}{ACC} & \multicolumn{1}{c}{F1} & \multicolumn{1}{c}{ACC} & \multicolumn{1}{c}{F1} \\
 \hline
\multirow{4}{*}{Trials} & Logistic & 86.49 & 88.49 & 84.64 & 81.63\\
 & SVM & 87.27 & 89.14 & 87.16 & 84.86 \\
 & RF & 82.66 & 85.59 & 90.94 & 89.96 \\
 & NN & \textbf{90.87} & \textbf{91.81} & \textbf{91.17} & \textbf{90.10} \\
 \hdashline
\multirow{4}{*}{\makecell{Participant ID}} & Logistic & 86.14 & 87.90 & 84.21 & 79.54 \\
 & SVM & 87.05 & 88.73 & 87.21 & 83.70 \\
 & RF & 82.99 & 85.57 & 89.92 & 87.56 \\
 & NN & \textbf{91.67} & \textbf{92.28} & \textbf{91.32} & \textbf{89.50} \\
 \hline
\end{tabular}
\end{center}
\end{table}

\subsubsection{Using Clustering to Improve Performance and Transferability}

We compare our clustering method with the approaches as listed in Sec.~\ref{sec:Using Cluster information to build Transferable Model}, and divide the dataset by participant ID.
Using clustering information achieves improvements on Japanese test data, while simply merging the data for joint training achieved the best average accuracy, as shown in Table~\ref{tab:transferable_model_for_zebra_crossing}.
One possible reason for this could be that, when the dataset is divided by participant ID, the NN model itself exhibits good transferability. Adding more data improves accuracy, while incorporating additional information may introduce noise, which could prevent further accuracy improvements.

\begin{table}[htb]
\caption{Results for transferable models for zebra crossing usage prediction. NN models are used. In percentage (\%). A higher score indicates better performance.}
\label{tab:transferable_model_for_zebra_crossing}
\begin{center}
\begin{tabular}{lrrrrrr}
\hline
 & \multicolumn{2}{c}{German Test} & \multicolumn{2}{c}{Japanese Test} & \multicolumn{2}{c}{Average} \\
 & \multicolumn{1}{c}{ACC} & \multicolumn{1}{c}{F1} & \multicolumn{1}{c}{ACC} & \multicolumn{1}{c}{F1} & \multicolumn{1}{c}{ACC} & \multicolumn{1}{c}{F1} \\
 \hline
Separate & 93.67 & 92.30 & 94.49 & 94.45 & 94.08 & 93.37 \\
Joint & \textbf{95.31} & \textbf{94.76} & 95.16 & 95.31 & \textbf{95.23} & \textbf{95.04} \\
Country & 93.92 & 92.95 & 94.26 & 94.41 & 94.09 & 93.68 \\
Cluster & 94.03 & 92.63 & \textbf{95.61} & \textbf{95.78} & 94.82 & 94.21 \\
\hline
\end{tabular}
\end{center}
\end{table}

\subsection{Trajectory Prediction}
\subsubsection{Comparison between Countries}
\label{sec: trajectory comparison}

We first use clustering to analyze pedestrian trajectories in each country.
When the number of clusters for each country is set to two, most trajectories naturally fall into two categories: those using zebra crossings and those not using zebra crossings. We compare the clustering results to the label of zebra crossing usage. The accuracy is notably high, achieving 97.4\% for data from Germany, and 99.4\% for data from Japan.
As the label of zebra crossing usage is determined based on the pedestrian's behavior when entering the road, the higher accuracy for participants in Japan indicates that they typically decide to use the zebra crossing before entering the road. Conversely, about 4.6\% of participants in Germany who initially did not use the zebra crossing changed their minds and decided to use it during the crossing process. This indicates differences in crossing behavior and decision-making processes between the two countries.

\begin{figure}[tb]
    \centering
    \includegraphics{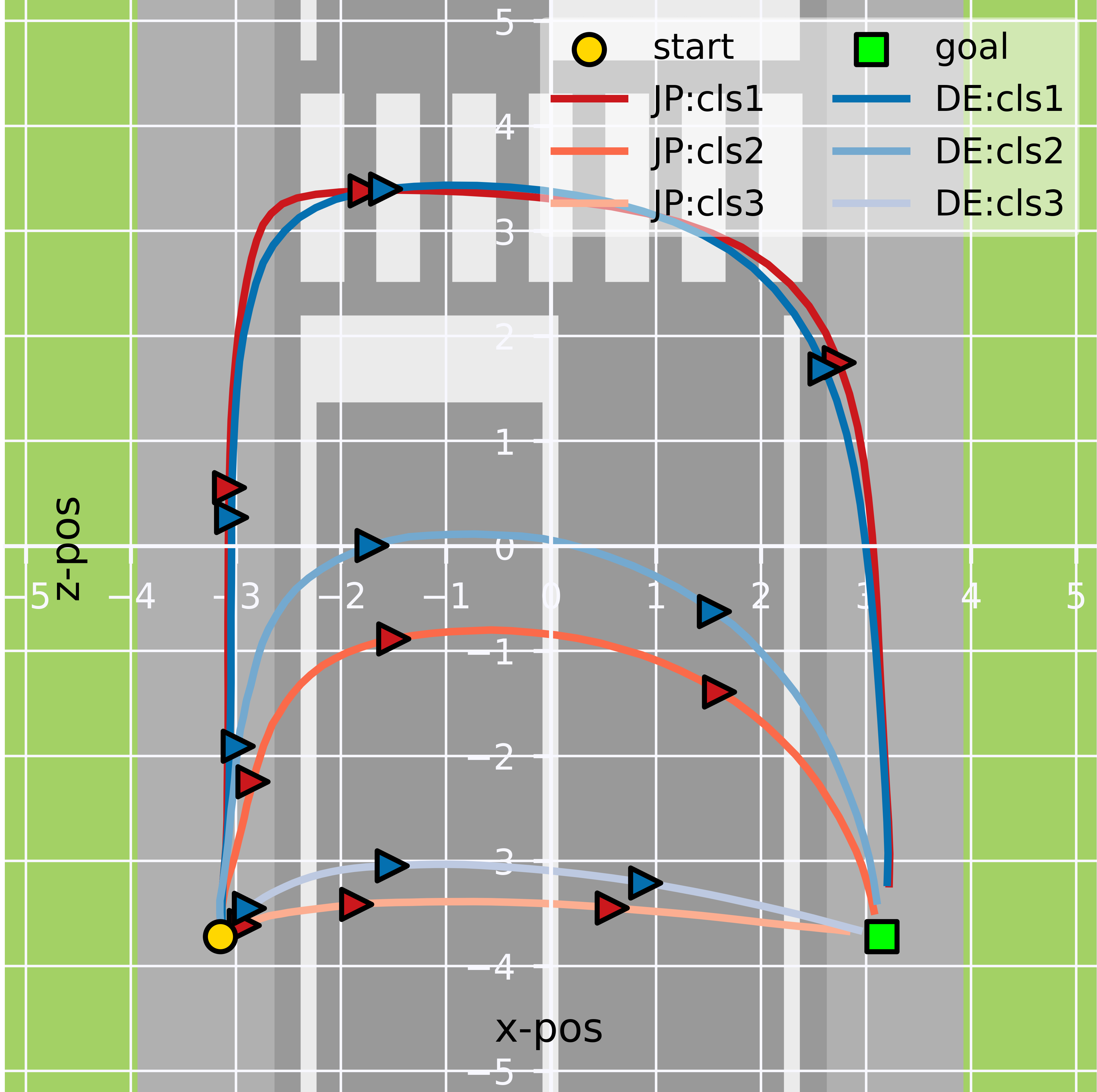}
    \caption{The average trajectories of the cluster (cls) analysis. JP stands for data collected in Japan, and DE stands for data collected in Germany. The 25\%, 50\%, and 75\% crossing duration is denoted by triangles on trajectories.}
    \label{fig:trajectories_clusters}
\end{figure}

To gain more detailed insights, we further clustered the trajectories into three groups. This resulted in three distinct categories: using zebra crossings (Cluster 1), direct crossing (Cluster 3), and trajectories in the middle (Cluster 2). The average trajectories of the three clusters for each country are plotted in Fig.~\ref{fig:trajectories_clusters}. The trajectories for pedestrians from the data collected in Japan using zebra crossings (Cluster 1) tend to be more upper left compared to those in Germany. This suggests that pedestrians from the data collected in Japan decide to use zebra crossings before starting to cross, while those from Germany might make this decision during the crossing. This is consistent with the clustering results, showing that pedestrians from data collected in Germany may change their minds after entering the road.
For pedestrians not using zebra crossings, those from data in Germany exhibit average trajectories closer to the zebra crossing, indicating they might consider using it but ultimately do not. In contrast, pedestrians from data collected in Japan display more direct crossing trajectories, showing less hesitation.

Comparing the trajectories of using and not using zebra crossings (Clusters 1 and 3) at the 25\% mark of the crossing duration, we observe different behavior patterns. Pedestrians using zebra crossings (Cluster 1) spend 25\% of their crossing time approaching the zebra marks, while pedestrians who cross directly (Cluster 3) are still before entering the road at the 25\% mark. This indicates that pedestrians who directly cross the road spend more time observing the traffic situation before starting to cross.

\begin{table*}[htb]
\caption{Average Displacement Error (ADE) for Trajectory Prediction of Pedestrians from the Data Collected in Germany and Japan, in meters (m).  A smaller error indicates better performance.}
\label{tab:ade_traj_german_and_japan}
\begin{center}
\begin{tabular}{cc|cccc|cccc|c}
\hline
&   & \multicolumn{4}{c|}{Germany}  & \multicolumn{4}{c|}{Japan}  \\
Features & Model  & \makecell[c]{Cluster 1 \\ (Zebra)} & \makecell[c]{Cluster 2 \\ (Middle)} & \makecell[c]{Cluster 3 \\ (Direct)} & All & \makecell[c]{Cluster 1 \\ (Zebra)} & \makecell[c]{Cluster 2 \\ (Middle)} & \makecell[c]{Cluster 3 \\ (Direct)} & All & Average \\ \hline
\multirow{3}{*}{\makecell[c]{Pre-event \\ input \\features}} & Linear & 1.003 & 0.886 & 0.867 & 0.935 & 1.157 & 0.720 & 1.419 & 1.240 & 1.088\\
& RF & \textbf{0.664} & \textbf{0.768} & 0.690  & \textbf{0.682} & 0.768 & \textbf{0.812} & 0.786 & 0.778 & 0.730\\
& NN & 0.669 & 0.926 & \textbf{0.679} & 0.688 & \textbf{0.626} & 1.075 & \textbf{0.683} & \textbf{0.673} & \textbf{0.681} \\ \hdashline

\multirow{3}{*}{\makecell[c]{With \\ additional \\features}} & Linear & 0.516 & 0.491 & 0.379 & 0.452 & 0.508 & 0.500 & 0.378 & 0.455 & 0.454 \\

& RF & \textbf{0.491} & \textbf{0.446} & \textbf{0.325} & \textbf{0.414} & \textbf{0.481} & \textbf{0.490} & \textbf{0.337} & \textbf{0.423} & \textbf{0.418} \\

& NN & 0.529 & 0.476 & 0.341 & 0.441 & 0.508 & 0.492 & 0.355 & 0.445 & 0.443 \\ \hline
\end{tabular}
\end{center}
\end{table*}

\begin{figure*}[htb]
\begin{center}

    \subfloat[German Cluster 1]{\includegraphics[width=0.3\textwidth]{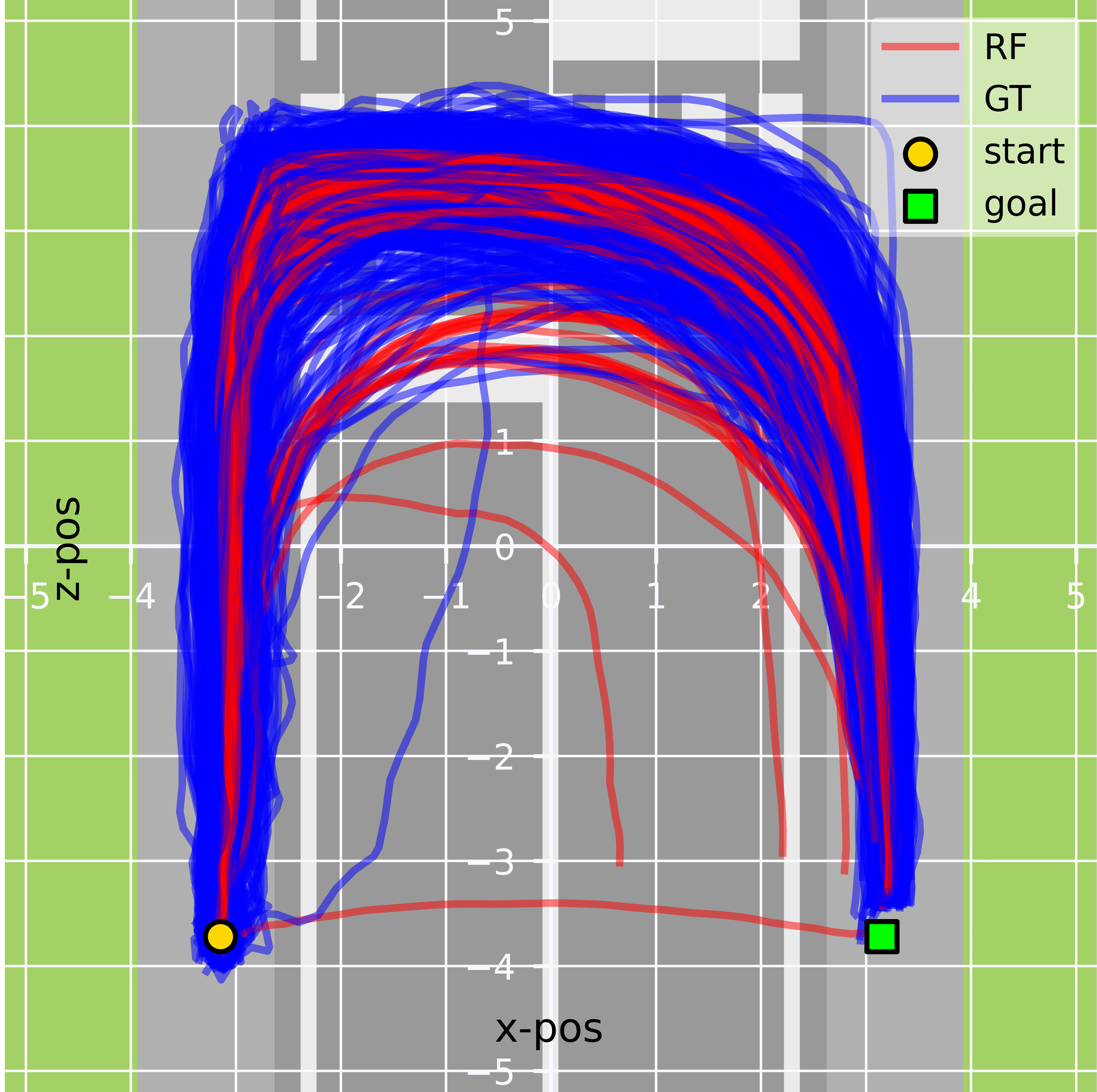}}
    \hfill
    \subfloat[German Cluster 2]{\includegraphics[width=0.3\textwidth]{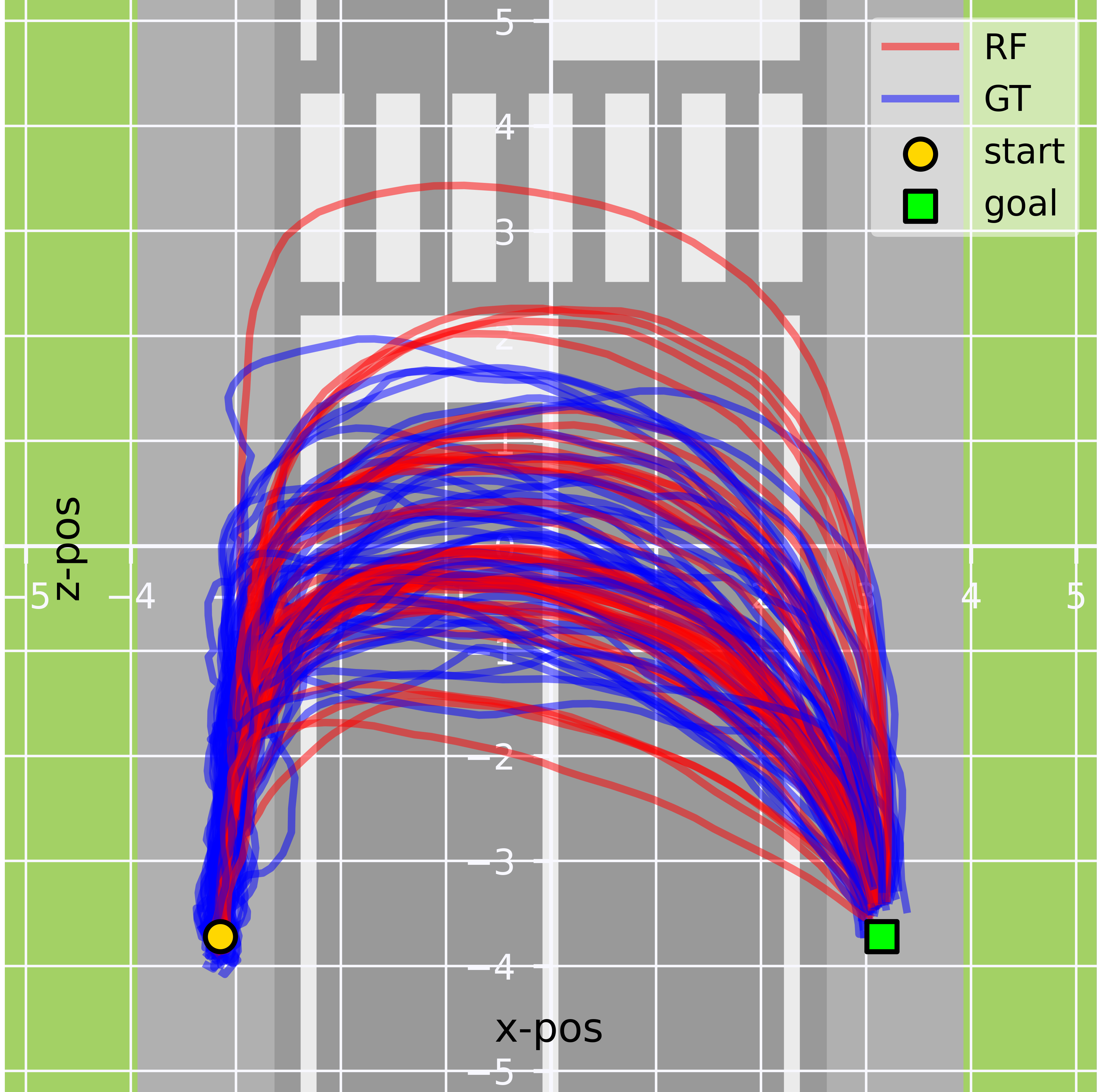}}
    \hfill
    \subfloat[German Cluster 3]{\includegraphics[width=0.3\textwidth]{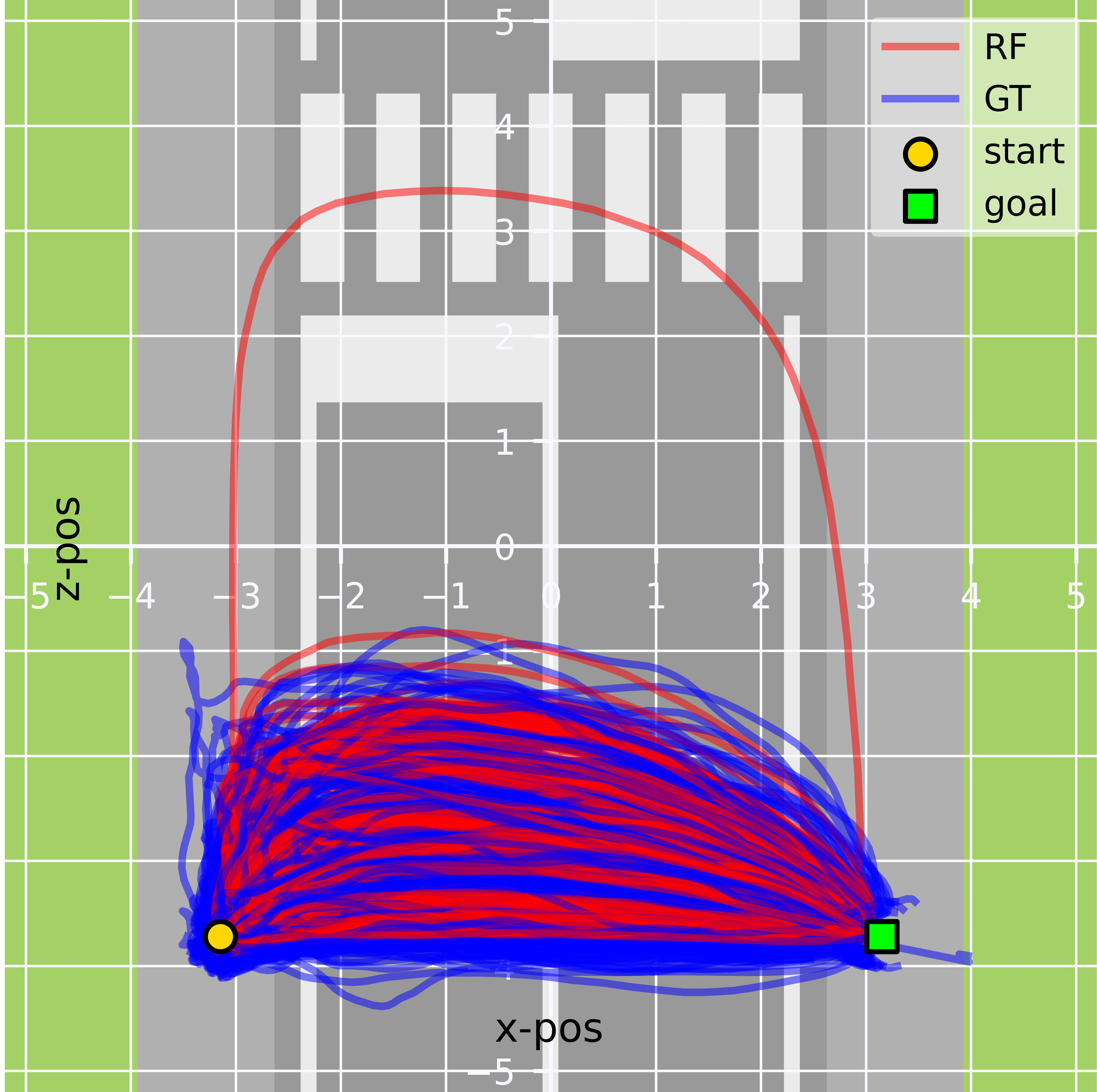}}
\vspace{0.2cm}
    \subfloat[Japanese Cluster 1]{\includegraphics[width=0.3\textwidth]{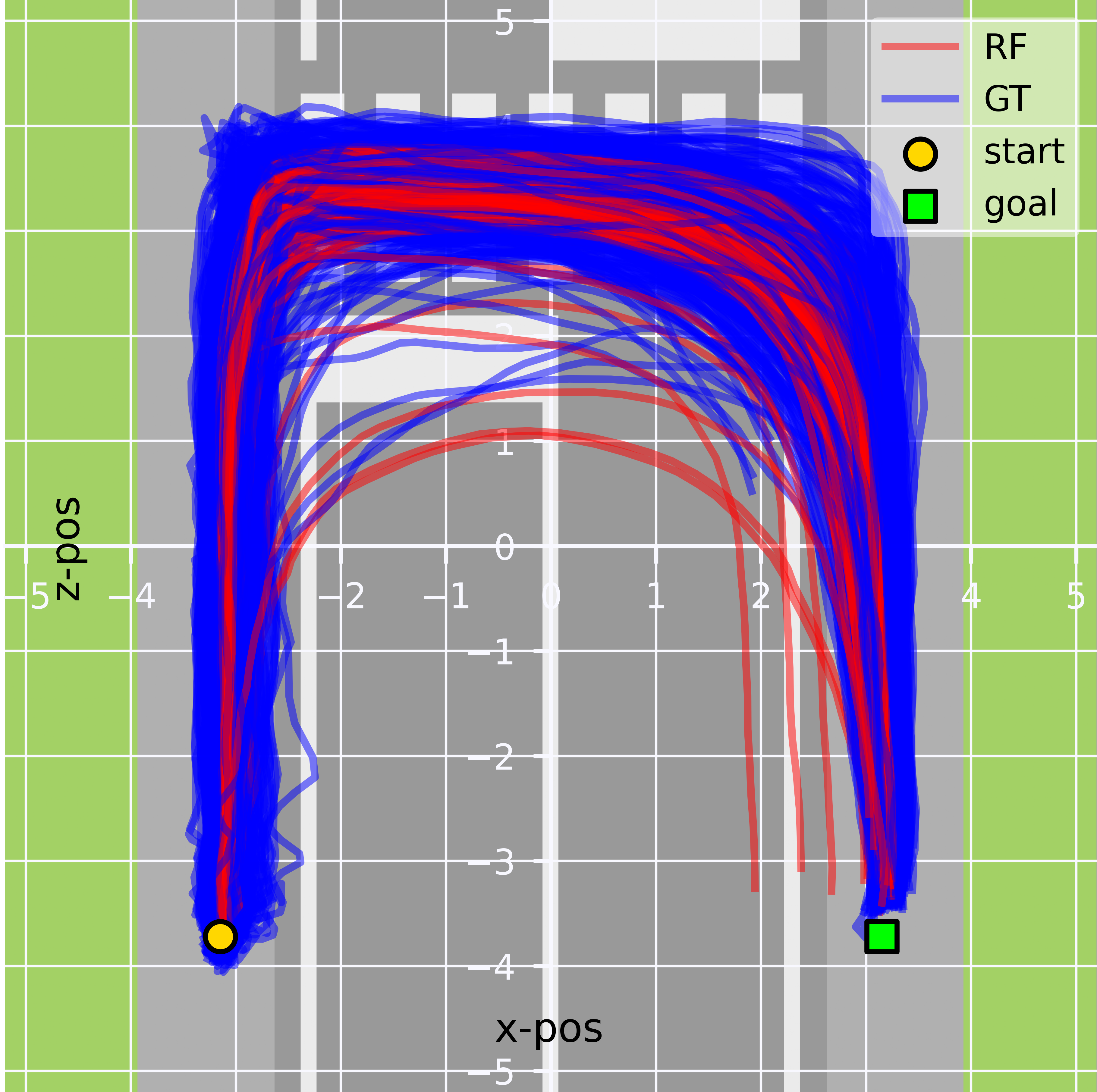}}
    \hfill
    \subfloat[Japanese Cluster 2]{\includegraphics[width=0.3\textwidth]{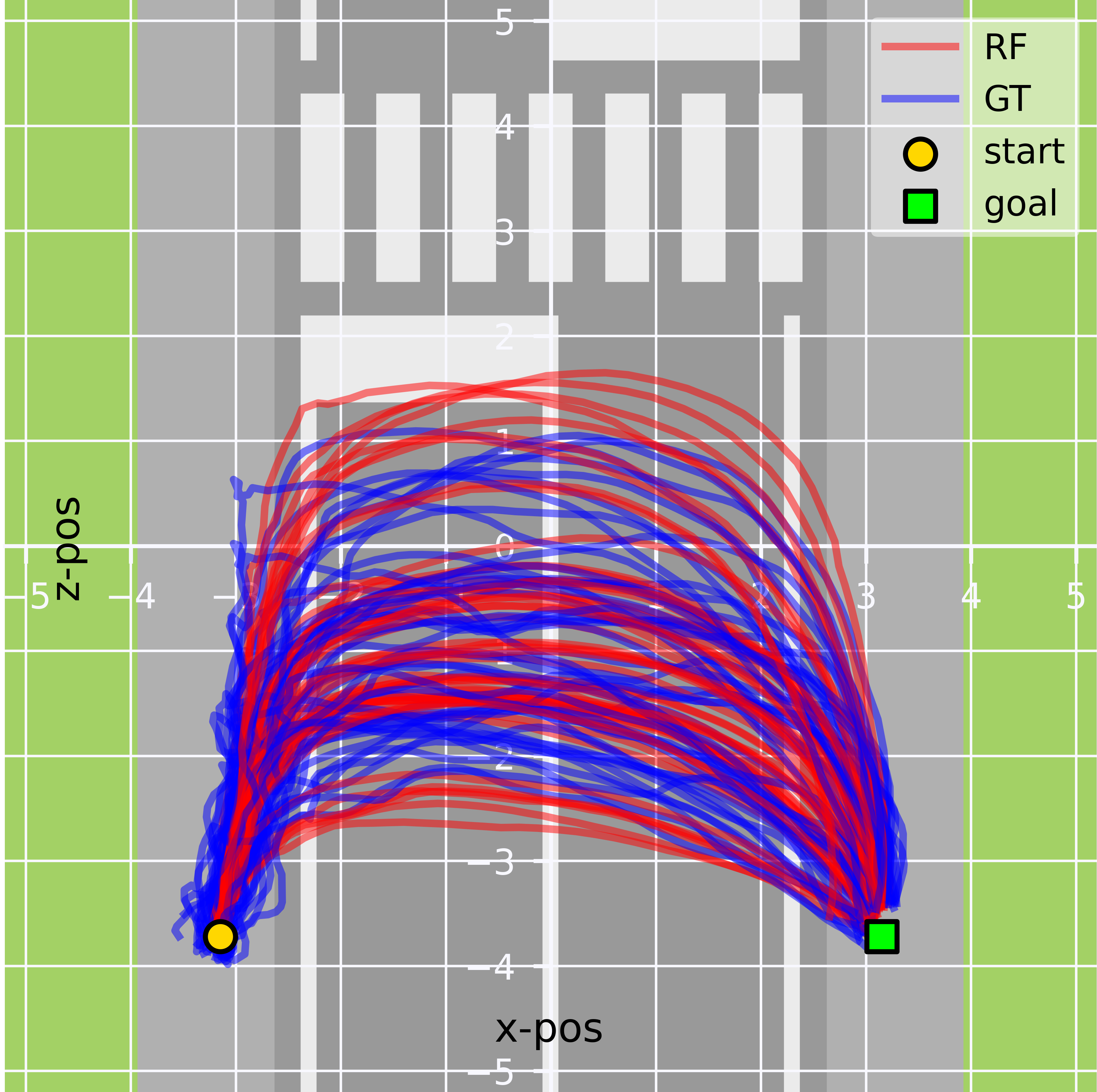}}
    \hfill
    \subfloat[Japanese Cluster 3]{\includegraphics[width=0.3\textwidth]{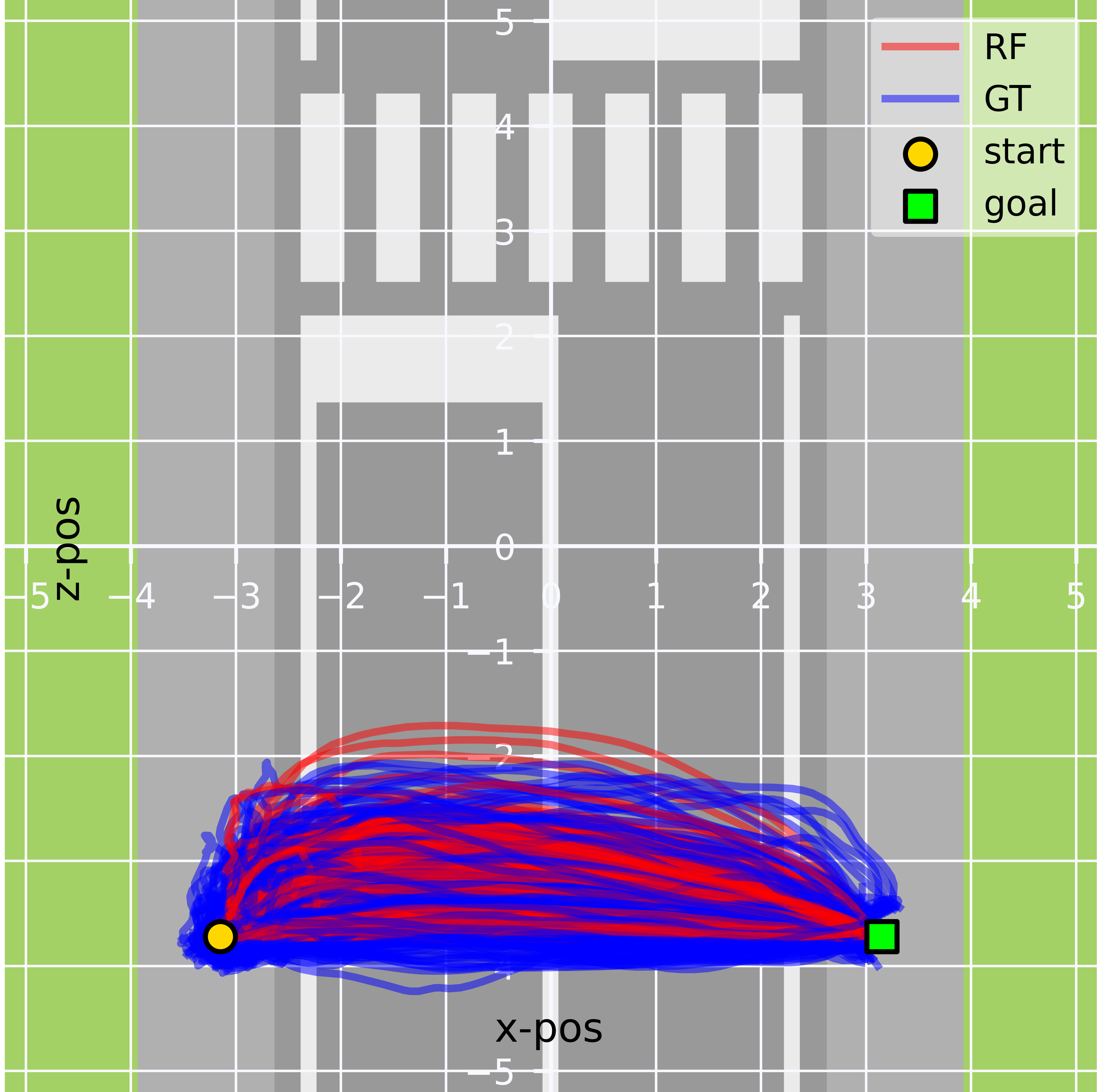}}
    \caption{Upper: (a)-(c) Trajectories illustrating RF model predictions compared to ground truth for participants in Germany. Lower: (d)-(e) Trajectories illustrating RF model predictions compared to ground truth for participants in Japan.}
    \label{fig:trajectory_prediction_results}
\end{center}
\end{figure*}

Pedestrians' trajectories are predicted and evaluated, with ADE shown in Table~\ref{tab:ade_traj_german_and_japan}. The NN model achieves the best performance with an average error of 0.681 m using only pre-event features. The RF model also shows effectiveness in specific clusters, especially for Cluster 2 in both countries. The linear regression model shows larger errors for data from Japan compared to data from Germany, while the NN model performs better on data from Japan than on data from Germany. This shows the non-linear nature of pedestrian trajectories for data from Japan.

Using RF and NN models, Clusters 1 and 3, corresponding to pedestrians either using zebra crossings or crossing directly, show lower errors compared to Cluster 2. This suggests higher predictability in decision-making for these clusters. Cluster 2, representing pedestrians who hesitate on whether to use the crossing, shows higher errors.

When additional features at the moment pedestrians enter the road are included, the RF model achieves the best results. The error significantly decreases compared to using only pre-event features. For participants in both countries, trajectories where pedestrians cross directly show smaller errors compared to trajectories in the other two clusters.

\begin{table*}[htb]
\caption{ADE of transferability in meters (m). A smaller error indicates better performance.}
\label{tab:trajectory_transfer_all}
\begin{center}
\begin{tabular}{c|cc|cccc|cccc|c}
\hline
& &   & \multicolumn{4}{c|}{Germany to Japan}  & \multicolumn{4}{c|}{Japan to Germany}  \\
Divided by & Features & Model  & \makecell[c]{Cluster 1 \\ (Zebra)} & \makecell[c]{Cluster 2 \\ (Middle)} & \makecell[c]{Cluster 3 \\ (Direct)} & All & \makecell[c]{Cluster 1 \\ (Zebra)} & \makecell[c]{Cluster 2 \\ (Middle)} & \makecell[c]{Cluster 3 \\ (Direct)} & All & Average \\  \hline

\multirow{6}{*}{Trials} & \multirow{3}{*}{\makecell[c]{Pre-event \\ input \\features}} & Linear & 1.112 & \textbf{0.783} & 2.137 & 1.510 & 1.496 & 0.941 & 1.321   & 1.386 & 1.448 \\
&  & RF & 0.703 & 0.954 & 1.418 & 1.007 & 0.980 & \textbf{0.887} & \textbf{0.595} & 0.801 & 0.904 \\
&  & NN & \textbf{0.634} & 0.887 & \textbf{1.378} & \textbf{0.950} & \textbf{0.743} & 1.158 & 0.621 & \textbf{0.711} & \textbf{0.831} \\ \cdashline{2-12}
  
& \multirow{3}{*}{\makecell[c]{With \\ additional \\features}} & Linear & 0.552 & \textbf{0.477} & 0.644 & 0.585 & 0.526 & 0.612 & 0.388 & 0.469 & 0.527 \\
&  & RF & \textbf{0.526} & 0.506 & \textbf{0.367} & \textbf{0.460} & \textbf{0.520} & \textbf{0.469} & \textbf{0.349} & \textbf{0.440} & \textbf{0.450} \\
&  & NN & 0.553 & 0.521 & 0.548 & 0.549 & 0.539 & 0.545 & 0.349 & 0.454 & 0.501 \\ \hline

\multirow{6}{*}{Participant ID} & \multirow{3}{*}{\makecell[c]{Pre-event \\ input \\features}} & Linear & 1.114 & \textbf{0.800} & 2.145 & 1.515 & 1.470 & 0.946 & 1.291 & 1.360 & 1.437 \\
& & RF & 0.679 & 1.025 & \textbf{1.450} & \textbf{1.010} & 0.938 & \textbf{1.010} & \textbf{0.596} & 0.788 & \textbf{0.899} \\
& & NN & \textbf{0.636} & 0.981 & 1.997 & 1.207 & \textbf{0.779} & 1.184 & 0.631 & \textbf{0.735} & 0.971 \\ \cdashline{2-12}
 
& \multirow{3}{*}{\makecell[c]{With \\ additional \\features}}  & Linear & 0.550 & \textbf{0.470} & 0.636 & 0.580 & 0.524 & 0.618 & 0.387 & 0.468 & 0.524 \\
& & RF & \textbf{0.525} & 0.534 & \textbf{0.383} & \textbf{0.468} & \textbf{0.522} & \textbf{0.475} & \textbf{0.341} & \textbf{0.438} & \textbf{0.453} \\
& & NN & 0.548 & 0.493 & 0.510 & 0.530 & 0.542 & 0.621 & 0.354 & 0.462 & 0.496 \\ \hline
 
\end{tabular}
\end{center}
\end{table*}

Fig.~\ref{fig:trajectory_prediction_results} compares the predictions of the RF model with ground truth trajectories for participants in Germany and Japan, using pre-event features (before entering the road) and additional features (at the moment when entering the road). The entire crossing trajectory was predicted using these inputs, without any trajectory information before the time point provided. The results are presented in three clusters. In Fig.~\ref{fig:trajectory_prediction_results} (a), there is an instance where the model incorrectly predicted a pedestrian would cross directly, while the ground truth shows the pedestrian initially considered crossing directly but then decided to use the zebra crossing. 
Accurately predicting such behavior with only pre-event information and additional features at the moment of entry is challenging. 
In Fig.~\ref{fig:trajectory_prediction_results} (c), another incorrectly predicted case occurs where the model predicted a pedestrian would use the zebra crossing, but the ground truth reveals that the pedestrian did not. This indicates the difficulty in predicting sudden changes in pedestrian behavior. Overall, while the trajectory predictions generally capture pedestrian motion accurately, they also highlight the complexities of predicting behaviors without using prior trajectory information.

\subsubsection{Model Transferability}
We evaluate the transferability of prediction models in Table~\ref{tab:trajectory_transfer_all} by performing transfer tasks. When the training and test sets are divided by trials, using only pre-event input features, the NN model performs the best. However, when additional features are included, the RF model achieves the smallest errors. The average error of the transfer tasks is larger compared to the non-transfer tasks. This could be attributed to the differences in data distributions between pedestrians from Germany and Japan.
For both using only pre-event features and with additional features, RF models achieve the best results, demonstrating their strong transferability.

\subsubsection{Using Clustering to Improve Performance and Transferability}

We first develop the model for trajectory prediction only using pre-event features. We explore strategies including separate training, joint training, and incorporating country information. We then explore three ways of grouping the data, leveraging cluster information from the training set to train a classifier aimed at reducing within-group disparities. The training and test sets are divided by participant IDs.

The results are shown in Table~\ref{tab:transferable_model_trajectory}.
It is indicated that joint training outperforms separate training approaches. Moreover, integrating country information further reduces prediction errors. Among the clustering methods tested, while grouping into two clusters reduces prediction error, grouping into three clusters further reduces errors, likely due to smaller within-group disparities.

Furthermore, leveraging zebra crossing usage labels to train a classifier and incorporating this information into model training achieves the best results, with an average prediction error of 0.632 m.

Finally, incorporating additional information when the pedestrian is entering the road improves performance compared to using only pre-event features. This enhancement results in the lowest prediction error of 0.432 m.

\begin{table}[htb]
\begin{center}
\caption{ADE for transferable models for trajectory prediction. NN and RF models are used. A smaller error indicates better performance.}
\label{tab:transferable_model_trajectory}
\begin{tabular}{c|cc|ccc}
\hline
Features & Strategy & Model & \makecell[c]{German \\Test} & \makecell{Japanese \\Test} & Average \\ \hline
\multirow{12}{*}{\makecell[c]{Pre-event \\ input \\features}}  & \multirow{2}{*}{Separate} & RF & 0.726 & 0.847 & 0.787 \\
 &  & NN & 0.739 & 0.837 & 0.788 \\
 & \multirow{2}{*}{Joint} & RF & 0.656 & 0.798 & 0.727 \\
 &  & NN & 0.687 & 0.814 & 0.751 \\
 & \multirow{2}{*}{Country} & RF & \textbf{0.658} & 0.798 & 0.728 \\
 &  & NN & 0.667 & \textbf{0.773} & \textbf{0.720} \\ \cdashline{2-6}
 & \multirow{2}{*}{\makecell[c]{Cluster \\(n=2)}} & RF & 0.625 & 0.704 & 0.664 \\
 &  & NN & 0.662 & 0.742 & 0.702 \\
 & \multirow{2}{*}{\makecell[c]{Cluster \\(n=3)}} & RF & 0.633 & 0.657 & 0.645 \\
 &  & NN & 0.661 & 0.694 & 0.678 \\
 & \multirow{2}{*}{Zebra Usage} & RF & \textbf{0.610} & \textbf{0.654} & \textbf{0.632} \\
 &  & NN & 0.680 & 0.718 & 0.699 \\ \hdashline
\multirow{4}{*}{\makecell[c]{With \\ additional \\features}} & \multirow{2}{*}{Separate} & RF & 0.427 & 0.465 & 0.446 \\
 &  & NN & 0.446 & 0.478 & 0.462 \\
 & \multirow{2}{*}{Joint} & RF & \textbf{0.416} & \textbf{0.449} & \textbf{0.432} \\
 &  & NN & 0.429 & 0.460 & 0.444 \\ 
\hline
\end{tabular}
\end{center}
\end{table}

\subsection{Implications for Intelligent Vehicles}

We focus on predicting and analyzing pedestrian behavior, including gap selection, zebra crossing usage, and crossing trajectories at unsignalized crossings. We compare differences between countries, evaluate model transferability, and propose a model using clustering information that performs better on data from both Germany and Japan. The proposed models have the potential to predict pedestrian behavior in advance and can be applied to smart traffic infrastructures and intelligent driving systems in different countries. 

This study employs symbolic data for prediction due to its interpretability, particularly when evaluating the importance of features. This improves transparency and understanding for users. The simplicity and efficiency of symbolic data facilitate lightweight real-time models that exhibit the potential to be deployed on vehicles, unrestricted by the source of the original data, such as camera images, LiDAR, radar, or from vehicular networks.

\subsection{Limitations}
This research is based on data collected from simulation environments. Although efforts were made to minimize the disparity between simulation data and real-world scenarios, a potential gap may remain that could influence the results. The time gap duration in these simulations was uniformly sampled between 2.5 s and 8.5 s, which differs from real-world distributions and may affect prediction accuracy.

\subsection{Future Work}

For future work, we can leverage generative models to generate additional pedestrian trajectories based on the collected data. More sophisticated models, such as LSTMs and Transformers can be employed for trajectory prediction and generation.
This can help enhance our understanding of the underlying data distribution of pedestrian behavior and capture complex behavior patterns. It has the potential to improve the robustness and generalizability of models.

\section{Conclusions}
In conclusion, this study advances the understanding and prediction of pedestrian behavior at unsignalized crossings. Our study has highlighted both the similarities and differences between pedestrians from Germany and Japan. Neural networks demonstrate better performance and transferability in predicting gap selection and zebra crossing usage. Random forests outperform other models in trajectory prediction. Pedestrians from the study conducted in Japan exhibit more cautious behavior and tend to select larger gaps.
The study also reveals that using unsupervised clustering methods can improve the accuracy of gap selection and trajectory prediction.
We have offered valuable insights into pedestrian behavior modeling and its transferability across different countries. The findings have practical implications for improving the safety and performance of smarter traffic by enabling more accurate and transferable predictive models.

\bibliographystyle{IEEEtran}
\bibliography{IEEEexample}

\begin{IEEEbiography}[{\includegraphics[width=1in,height=1.25in,clip,keepaspectratio]{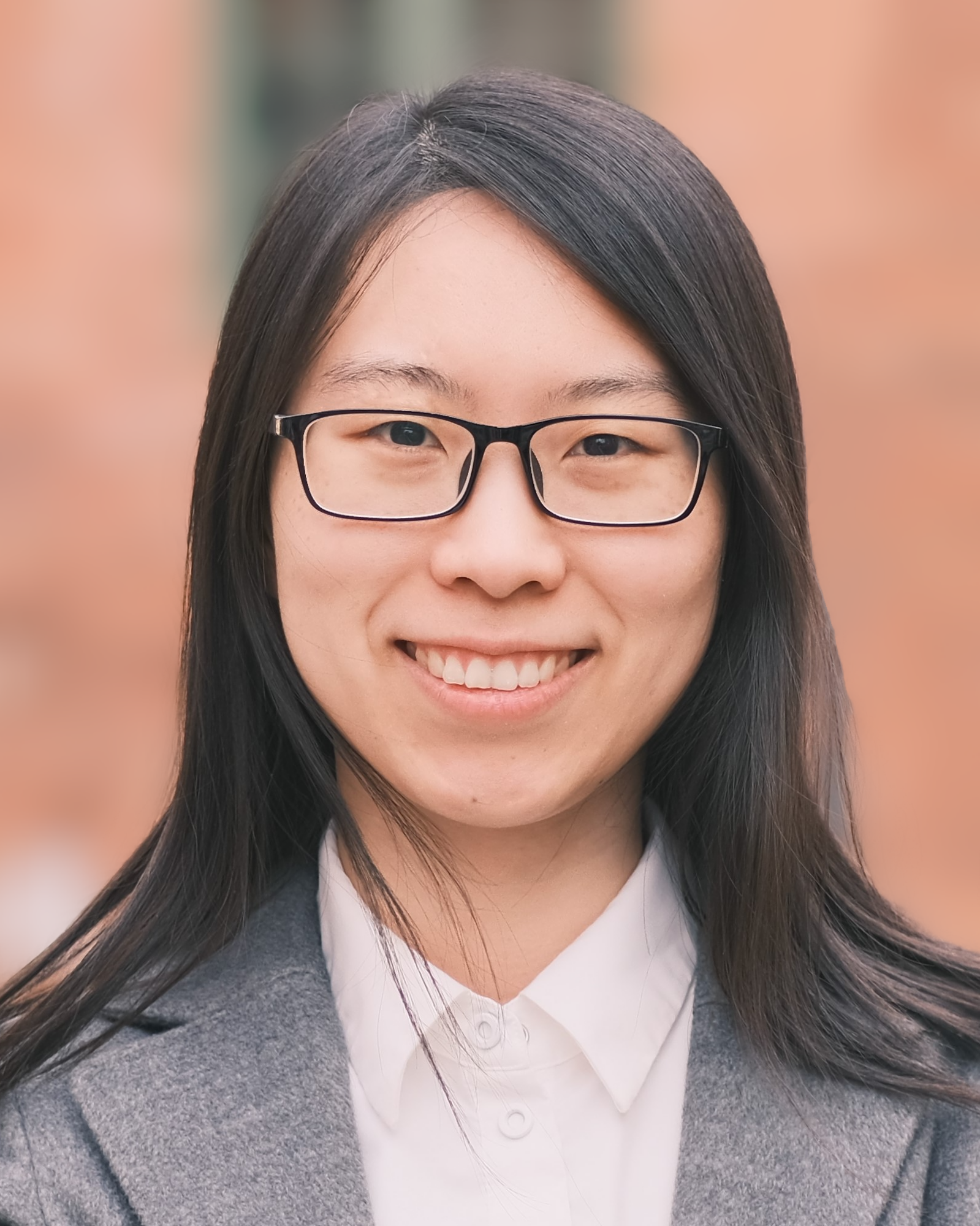}}]{Chi Zhang}
received the B.Eng. and M.Eng. degrees in Control Science and Engineering from Zhejiang University, China, in 2014 and 2017, respectively, and the Licentiate of Philosophy degree in Computer Science and Engineering from the University of Gothenburg, Sweden, in 2022. Between 2017 and 2020, she was a research and development engineer at the Intelligence Driving Group, Baidu, China, in the area of automated driving perception. Since 2020, she has been pursuing a Ph.D. degree with the Department of Computer Science and Engineering, University of Gothenburg, Sweden. She is a Marie Curie Early Stage Researcher, and an affiliated Ph.D student of the Wallenberg AI, Autonomous Systems and Software Program (WASP) program.
Her research interests include deep learning, machine learning, computer vision, road user behavior prediction, perceptions for automated driving, and robotics.
\end{IEEEbiography}

\begin{IEEEbiography}[{\includegraphics[trim={1in 0 1in 0}, width=1in,height=1.25in,clip,keepaspectratio]{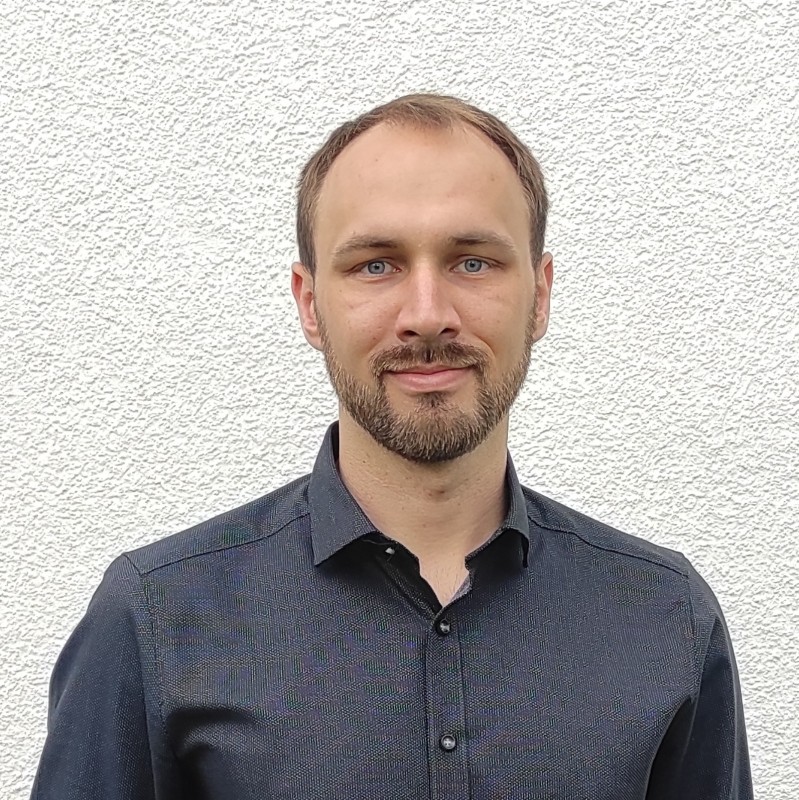}}]{Janis Sprenger} received the B.Sc. in Computer Science, the M.Sc. in Computer Science, and the B.Sc. in Psychology from Saarland University, Germany, in 2016, 2019, and 2019, respectively. Between 2018 and 2023, he worked as a researcher at the German Research Center for Artificial Intelligence (DFKI). Since 2023, he is leading a team on Motion Intelligence and Agents at DFKI. Since 2019, he is pursuing a Ph.D. degree with the Department of Computer Science at Saarland University, Germany. His research interests include generative AI, digital twins, motion capturing, motion analysis, and data-driven motion synthesis. 
\end{IEEEbiography}

\begin{IEEEbiography}[{\includegraphics[width=1in,height=1.25in,clip,keepaspectratio]{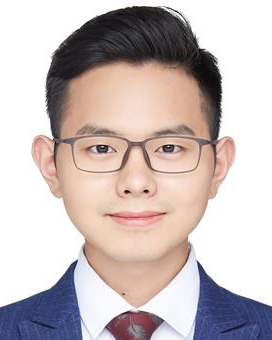}}]{Zhongjun Ni}
received the B.Eng. and M.Eng. degrees from Zhejiang University, China, in 2014 and 2017, respectively, and the Licentiate of Engineering degree from Link\"oping University, Sweden, in 2023. Between 2017 and 2020, he worked as a software engineer in the industry, such as Microsoft. He is currently pursuing his Ph.D. degree with the Department of Science and Technology at Link\"oping University, Sweden. His research interests include time series analysis, digital twins, and Internet of Things solutions based on Edge-Cloud computing.
\end{IEEEbiography}

\begin{IEEEbiography}[{\includegraphics[width=1in,height=1.25in,clip,keepaspectratio]{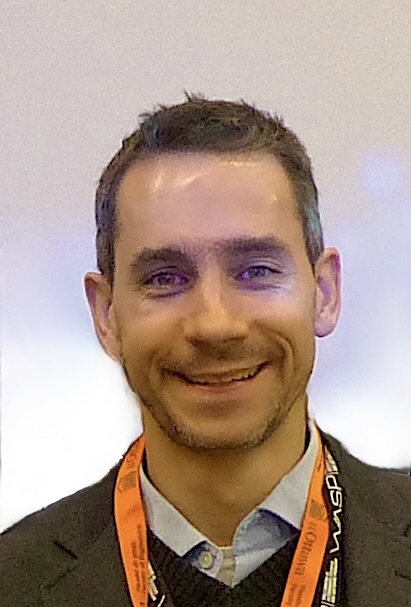}}]{Christian Berger}
Dr.~Christian Berger is Full Professor at the Department of Computer Science and Engineering at University of Gothenburg, Sweden and received his Ph.D.~degree from RWTH Aachen University, Germany in 2010. He coordinated the research project for the self-driving vehicle ``Caroline'', which participated in the 2007 DARPA Urban Challenge Final--the world's first urban robot race. He co-led the Chalmers Truck Team during the 2016 Grand Cooperative Driving Challenge (GCDC) and is the leading software architect at Chalmers Revere, the laboratory for automotive-related research. His research expertise is on architecting cloud-enabled cyber-physical systems and event identification in multi-modal, large-scale, time-series datasets.
\end{IEEEbiography}

\vfill

\end{document}